\documentclass[runningheads]{llncs}

\usepackage{eccv}

\usepackage{eccvabbrv}

\usepackage{graphicx}
\usepackage{booktabs}
\usepackage{algorithm,algorithmic}

\usepackage{verbatim}
\usepackage{tabularx}
\usepackage{xspace}
\usepackage{bbm}
\usepackage{bm}
\usepackage{booktabs}
\usepackage{amssymb}
\usepackage{mathtools}
\usepackage{wrapfig}

\usepackage{amssymb,amsmath}
\usepackage{nicematrix}

\usepackage{colortbl}

\usepackage{multirow}
\usepackage{booktabs}

\usepackage{amssymb}
\usepackage{pifont}

\definecolor{mygreen}{RGB}{20, 180, 70}
\definecolor{myorange}{RGB}{255, 140, 0}
\definecolor{myred}{RGB}{220, 30, 30}

\newcommand{\cmark}{{\color{mygreen}\ding{51}}}
\newcommand{\cmarkOrange}{{\color{myorange}(\ding{51})}}
\newcommand{\xmark}{{\color{myred}\ding{56}}}

\newlength{\smallimage}
        \definecolor{rel}{rgb}{.1,.6,.2}
        \definecolor{nrl}{rgb}{1,1,1}
        \definecolor{qim}{rgb}{1,1,1}

\makeatletter

\def\eg{\emph{e.g.},\,}

\def\ie{\emph{i.e.},\,}

\def\etc{\emph{etc.\,}}
\def\vs{\emph{vs.\,}}
\def\wrt{w.r.t.\,}

\def\etal{\emph{et al.\,}}

\def\be{\begin{equation}}
\def\ee{\end{equation}}
\def\bea{\begin{eqnarray}}
\def\eea{\end{eqnarray}}
\def\ben{\begin{eqnarray*}}
\def\een{\end{eqnarray*}}

\def\bi{\begin{itemize}}
\def\ei{\end{itemize}}

\newcommand{\btab}[1]{\begin{tabular}{#1}}
\newcommand{\etab}{\end{tabular}}
\newcommand{\ba}[1]{\begin{array}{#1}}
\newcommand{\ea}{\end{array}}

\def\<{\langle}
\def\>{\rangle}

\newcommand{\myparagraph}[1]{\vspace{0.2cm}\noindent\textbf{#1.}}


\definecolor{sb_blue}{rgb}{0.2,0.45,0.63}
\definecolor{sb_red}{rgb}{0.84,0.16,0.16}
\definecolor{sb_violet}{rgb}{0.58,0.4,0.74}
\definecolor{sb_orange}{rgb}{0.9,0.5,0.2}
\definecolor{sb_green}{rgb}{0.17,0.63,0.17}
\definecolor{sb_brown}{rgb}{0.55,0.34,0.29}

\definecolor{DarkCoral}{rgb}{0.8, 0.36, 0.27}

\usepackage[accsupp]{axessibility}  

\usepackage{hyperref}
\usepackage[capitalize]{cleveref}

\creflabelformat{equation}{#2\textup{(\textcolor{red}{#1})}#3}
\creflabelformat{table}{#2\textcolor{red}{#1}#3}
\creflabelformat{figure}{#2\textcolor{red}{#1}#3}
\creflabelformat{section}{#2\textcolor{red}{#1}#3}
\creflabelformat{appendix}{#2\textcolor{red}{#1}#3}
\creflabelformat{algorithm}{#2\textcolor{red}{#1}#3}

\Crefname{equation}{Eq.}{Eqs.}
\Crefname{figure}{Fig.}{Figs.}
\Crefname{tabular}{Tab.}{Tabs.}
\Crefname{table}{Tab.}{Tabs.}
\Crefname{section}{Sec.}{Secs.}

\crefname{equation}{Eq.}{Eqs.}
\crefname{figure}{Fig.}{Figs.}
\crefname{tabular}{Tab.}{Tabs.}
\crefname{table}{Tab.}{Tabs.}
\crefname{section}{Sec.}{Secs.}

\usepackage{orcidlink}

\begin{document}

\title{Placing Objects in Context via Inpainting for Out-of-distribution Segmentation} 

\titlerunning{Placing Objects in Context via Inpainting}

\author{Pau de Jorge\inst{1} 
\and
Riccardo Volpi\inst{1} \and
Puneet K. Dokania \inst{2}
\and \\ Philip H.S. Torr \inst{2} \and Gr\'egory Rogez \inst{1}}

\authorrunning{de Jorge et al.}

\institute{
Naver Labs Europe \url{https://europe.naverlabs.com/} \and 
University of Oxford \\
\email{pau.dejorge@naverlabs.com} \\
}

\maketitle

\setlength{\intextsep}{5pt}%
\setlength{\columnsep}{10pt}%

\begin{abstract}

When deploying a semantic segmentation model into the real world, it will inevitably encounter semantic classes that were not seen during training. To ensure a safe deployment of such systems, it is crucial to accurately evaluate and improve their \emph{anomaly segmentation} capabilities. However, acquiring and labelling semantic segmentation data is expensive and unanticipated conditions are long-tail and potentially hazardous. Indeed, existing anomaly segmentation datasets capture a limited number of anomalies, lack realism or have strong domain shifts. In this paper, we propose the Placing Objects in Context (POC) pipeline to realistically add \emph{any} object into \emph{any} image via diffusion models. POC can be used to easily extend any dataset with an arbitrary number of objects. In our experiments, we present different anomaly segmentation datasets based on POC-generated data and show that POC can improve the performance of recent state-of-the-art anomaly fine-tuning methods across several standardized benchmarks. POC is also effective for learning new classes. For example, we utilize it to augment Cityscapes samples by incorporating a subset of Pascal classes and demonstrate that models trained on such data achieve comparable performance to the Pascal-trained baseline. This corroborates the low synth2real gap of models trained on POC-generated images. Code: \url{https://github.com/naver/poc}
\keywords{Anomaly segmentation \and OOD segmentation \and Inpainting}
\end{abstract}    
\section{Introduction}
\label{sec:intro}

When 
we deploy
autonomous agents such as robots or self-driving cars,
we expose them
to the unpredictable nature of the real world. Inevitably, they will encounter visual conditions that were not anticipated during training. 
In particular, the presence of unseen objects in the scene poses a significant safety hazard. For example, consider an unknown wild animal crossing the street and being classified by the model as ``road''. 
To tackle this issue, it is important to distinguish out-of-distribution (OOD) categories---\ie novel objects unseen during training---from the in-distribution (ID) ones. In the context of semantic image segmentation, this task is often referred to as \textit{anomaly segmentation}.

Although several methods have been proposed that allow for segmenting anomalies~\cite{grcic2022densehybrid, lis2020detecting, jung2021standardized, liang2022gmmseg, di2021pixel, besnier2021triggering}, 
accurately evaluating the performance of such methods is 
a challenge in itself.
Given a model trained on a particular dataset, 
the aim is to test its ability to distinguish OOD categories in conditions that resemble the training domain.
For example, to test a model trained for semantic
segmentation of urban scenes, 
the ideal test set would be constituted by images showing
OOD categories within an urban environment similar
to the training one.
Yet, 
potential anomalies 
follow a long-tailed distribution
and it is inefficient, or even hazardous, to acquire and label images with 
arbitrary OOD objects.

\begin{figure}[t]
  \begin{minipage}[b]{.5\linewidth}
        \centering
        \includegraphics[width=0.99\linewidth]{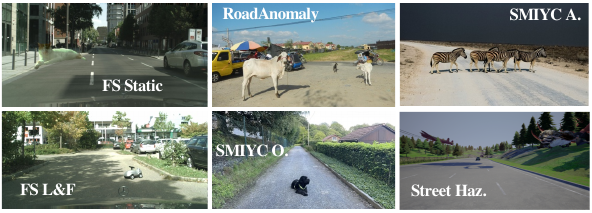}
        \captionof{figure}{
        \small \textbf{Samples from previous OOD datasets.} FS Static has unrealistic OOD objects while RoadAnomaly and SMIYC datasets have strong domain shifts from Cityscapes. FS L\&F (which manually inserts OOD objects) and StreetHazards (full simulation) have large set-up costs.
        }
        \label{fig:dset_comparison}
  \end{minipage}\hfill
  \begin{minipage}[b]{.45\linewidth}
        \scriptsize
        \centering
        \begin{tabular}{lcccc}
        \toprule
        {} &
          \textbf{No} &
          \textbf{OOD} &
          {} &
          {\textbf{Low}} \\ 
          \multirow{-2}{*}{\textbf{Dataset}} &
          \textbf{shift} &
          \textbf{realism} &
           \multirow{-2}{*}{\textbf{Dynamic}} &
           \textbf{cost} \\ 
        \midrule
        FS Static~\cite{blum2019fishyscapes} &
          \cmark &
          \xmark &
          \cmarkOrange &
          \cmarkOrange \\
        FS L\&F~\cite{pinggera2016lost} &
          \cmarkOrange &
          \cmark &
          \xmark &
          \xmark \\
        SMIYC O.~\cite{chan2021segmentmeifyoucan} &
          \xmark &
          \cmark &
          \xmark &
          \xmark \\
        SMIYC A.~\cite{chan2021segmentmeifyoucan} &
          \xmark &
          \cmark &
          \xmark &
          \xmark \\
        RoadAn.~\cite{lis2019detecting} &
          \xmark &
          \cmark &
          \xmark &
          \xmark \\
        StreetHaz.~\cite{hendrycks2019scaling} &
          \xmark &
          \cmarkOrange &
          \cmarkOrange &
          \xmark \\
        POC (ours) &
          \cmark &
          \cmarkOrange &
          \cmark &
          \cmark \\
        
        \bottomrule
        \end{tabular}
        \captionof{table}{\small \textbf{Comparison of anomaly test sets.} We qualitatively compare datasets on four main axes. We score them as either good (\cmark), medium (\cmarkOrange) or bad (\xmark). Further discussion in \cref{sec:ood_sets}.}
        \label{tab:dset_comparison}
  \end{minipage}
\end{figure}

Previous approaches to generating anomaly segmentation datasets can be grouped into three families: \textit{Stitching and blending} OOD objects from other sources into images from 
the
original
dataset \cite{blum2019fishyscapes}; \textit{Collecting images} from driving scenes 
and annotating
OOD objects \cite{pinggera2016lost, chan2021segmentmeifyoucan, lis2019detecting}; \textit{Full simulation} of urban scenes 
with
anomalies~\cite{hendrycks2019scaling}. \textit{Stitching and blending} is relatively 
inexpensive 
if
OOD objects are segmented elsewhere, yet it often leads to unrealistic insertions (\eg object's size or illumination). \textit{Collected images} 
contain real anomalies, but 
are expensive to acquire
and have 
a significant
distribution shift 
from the original dataset (it is not easy to find or replicate images following the original setup).
\textit{Full simulation} allows for perfectly blended objects but 
bears
a high setup cost and results in severe drifts from  the training distribution. See examples in \cref{fig:dset_comparison}.

Ideally, methods to generate anomaly segmentation datasets should satisfy four main desiderata: \textit{i)} 
Minimal domain shift with respect to the training set---since large domain shifts may lead to underestimating anomaly segmentation capabilities; \textit{ii)} 
Generating realistic images; \textit{iii)} 
Allowing for a dynamic
generation of new images with arbitrary OOD objects; \textit{iv)} 
Incurring low setup costs.

This motivates us to 
introduce
the \textbf{Placing Objects in Context} pipeline (POC),
which enables practitioners to
generate anomaly segmentation test sets by realistically inserting \textit{any} object (OOD or ID) into \textit{any} image on the fly.
See a comparison of approaches followed to generate previous benchmarks and the proposed POC along our desiderata  in~\cref{tab:dset_comparison}.


To realistically insert new objects, we control the location of the added object and apply only local changes to preserve the overall scene semantics. We utilize open-vocabulary segmentation~\cite{Grounded-SAM_Contributors_Grounded-Segment-Anything_2023} to select valid regions where the object can be placed, such as ``the road''. We then feed the selected region to an inpainting model~\cite{rombach2022high} with a conditioning prompt, such as ``a cat''. After inpainting, we apply again the segmentation model to the modified area to automatically annotate the added object and detect generation failures, \ie when the object was not properly generated.



\begin{figure}[t]
\centering
\includegraphics[width=\linewidth]{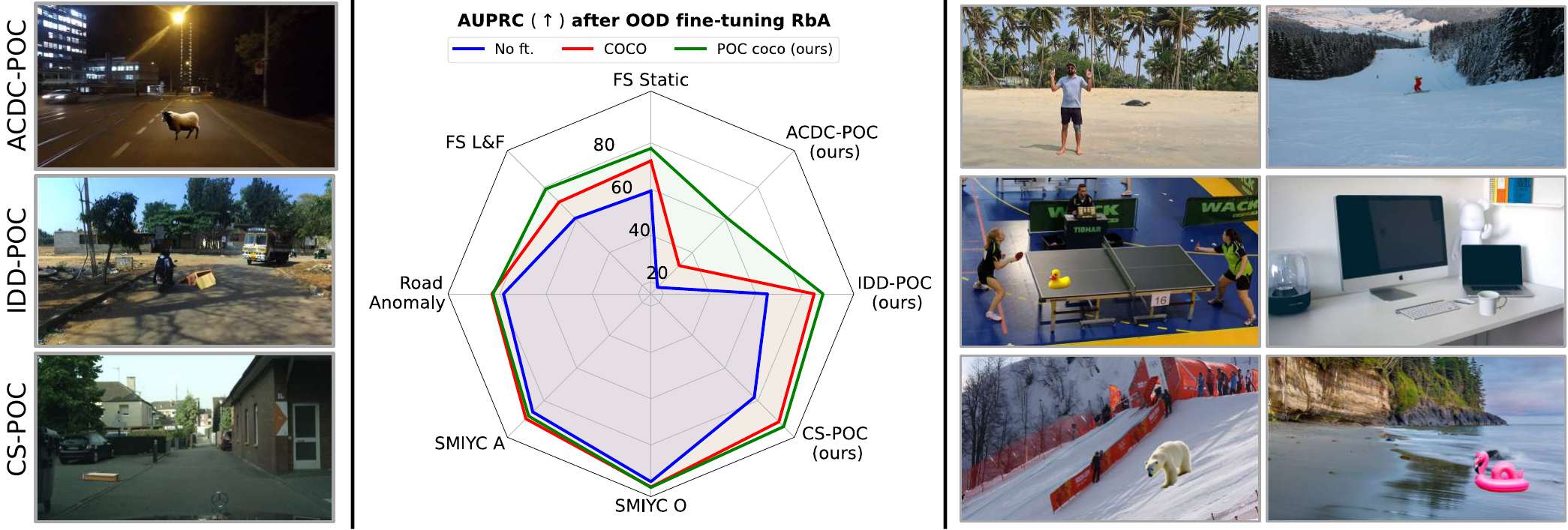} 

\caption{\small \textbf{Left:} Samples of our POC-generated datasets. Top to bottom, inserted anomalies are ``sheep'', ``dumped furniture'' and ``carton box''. \textbf{Middle:} AUPRC on different anomaly segmentation datasets. We evaluate RbA~\cite{nayal2023rba} prior to fine-tuning, and after fine-tuning with COCO objects or POC-generated images. Fine-tuning with POC improves results on several benchmarks. \textbf{Right:} Beyond road scenes, POC can be applied seamlessly in diverse scenes. Clockwise, inserted objects are: ``sea turtle'', ``person skiing'', ``white porcelain mug'', ``inflatable flamingo'', ``polar bear'' and ``rubber duck''.
}
\label{fig:splash}
\end{figure}

In our experiments, we show that fine-tuning on POC-generated data can significantly improve the performance of state-of-the-art anomaly segmentation methods---outperforming models fine-tuned via the standard practice of \textit{stitching} COCO objects. 
We also present three POC-generated evaluation sets based 
on urban scene segmentation datasets~\cite{cordts2016cityscapes,sakaridis2021acdc,varma2019idd},
and benchmark different anomaly segmentation methods on them (see~\cref{fig:splash} (middle) for a first glimpse of results).

Finally, since POC can add arbitrary objects, we show it can be used to learn new classes. For instance, augmenting Cityscapes~\cite{cordts2016cityscapes} images with animal classes leads to $93.14$ mIoU on Pascal's test set~\cite{pascal-voc-2012} 
(using
the same classes) \textit{without seeing any real animal}, while directly training on Pascal yields $94.75$---namely, models trained on POC-edited images exhibit a rather small synth2real gap. 
%
%

\section{Related work}
\label{sec:rel_work}

We cover the relevant literature on methods and datasets for anomaly segmentation and on diffusion models---the research areas most related to our work.

\myparagraph{Anomaly segmentation methods}
Early
works relied on approximating uncertainty via softmax probabilities \cite{hendrycks2016baseline, liang2017enhancing}, model ensembles \cite{lakshminarayanan2017simple} or dropout \cite{mukhoti2018evaluating, gal2016dropout}. Yet, models tend to be overconfident, resulting in high confidence also for OOD samples \cite{nguyen2015deep, guo2017calibration, jiang2018trust}. Alternative confidence measures have been proposed 
that either rely on logits \cite{corbiere2019addressing, hendrycks2019scaling, jung2021standardized, liu2020energy} or density estimators \cite{lee2018simple}. 
Another body of works reconstruct images with
generative models and detect anomalies as discrepancies between original images and their reconstructions~ \cite{haldimann2019not, lis2019detecting, lis2020detecting, xia2020synthesize}. 
Currently, the most promising methods use OOD data to fine-tune the models
\cite{chan2021entropy, tian2022pixel, grcic2022densehybrid}. In particular, they crop OOD objects from COCO \cite{lin2014microsoft} and \textit{stitch} them in Cityscapes images.
In this work, we
build on
three state-of-the-art OOD fine-tuning methods \cite{nayal2023rba, liu2023residual, rai2023unmasking}
and combine them with POC.

\myparagraph{Anomaly segmentation datasets}
\label{sec:ood_sets}
We compare existing anomaly segmentation datasets across four axes (\textit{Domain shift}, \textit{OOD realism}, \textit{Dynamism} and \textit{Set up cost})---see~\cref{tab:dset_comparison} for a summary and~\cref{fig:dset_comparison} for sample images.
The \textbf{Fishyscapes Static}~\cite{blum2019fishyscapes} approach involves randomly stitching OOD objects from Pascal \cite{pascal-voc-2012} onto Cityscapes images. 
While this method avoids domain shift, the stitched OOD objects lack realism.
If OOD object images and masks are available,
datasets can be generated \textit{dynamically}. However, if new objects are required, they must be obtained from additional datasets or from the web incurring moderate \textit{setup costs}.
At the other end of the spectrum, \textbf{RoadAnomaly} \cite{lis2019detecting} and \textbf{Segment-me-if-you-can (SMIYC)} datasets \cite{chan2021segmentmeifyoucan} contain real images with anomalies downloaded from the web. While this ensures \textit{OOD realism}, it often leads to a large \textit{domain shift}. Moreover, manual labelling of OOD objects has a significant \textit{setup cost} and is \textit{not dynamic}, \ie new images would need to be acquired and labelled to generate new samples. \textbf{Lost \& Found} \cite{pinggera2016lost} mimics the Cityscapes setup to reduce domain shift, but OOD objects have been inserted artificially, leading to low variability and difficulty in scaling. 
\textbf{Street Hazards} \cite{hendrycks2019scaling} shows a fully simulated dataset that allows for \textit{dynamic} generation of images while the simulation engine inserts OOD objects \textit{realistically} in terms of lighting. Yet, pose and size of the object are pseudo-random, which is not always realistic and leads to a strong \textit{domain shift} from simulation to real images. Moreover, it requires an accurate 3D model of all objects, which bears a significant \textit{setup cost} and hinders scalability.

In contrast to prior art, our proposed pipeline 
is plug-and-play and allows for adding objects into images with \textit{no setup costs}. Built on top of open-vocabulary models, it can \textit{dynamically} insert any object by changing the text prompts. We observe both qualitatively and quantitatively that our pipeline leads to greater \textit{OOD realism} and applying inpainting also helps to mitigate \textit{domain shift}.\\

\myparagraph{Diffusion models} 
Introduced
by Sohl-Dickstein~\etal \cite{sohl2015deep}, diffusion models have led to unprecedented quality in image generation \cite{dhariwal2021diffusion, ho2020denoising, saharia2022palette, song2019generative}. In particular, text-to-image models 
condition the image generation or image editing process
on a given text prompt
\cite{nichol2021glide, ramesh2022hierarchical, rombach2022high, saharia2022photorealistic}. 
%
%
Image inpainting methods, which
insert
objects locally by only modifying masked regions of an image~\cite{rombach2022high, avrahami2022blended, ramesh2022hierarchical}, are particularly relevant to our goal. 
General-purpose editing models that can edit images based on text prompts \cite{brooks2023instructpix2pix, meng2021sdedit} 
are also 
related to our work.
Among the latter,
InstructPix2Pix \cite{brooks2023instructpix2pix} has recently shown very realistic results following text instructions, \eg ``make the photo look like it was taken at sunset''. Yet, we found that it often fails to add new objects to the scene \eg ``add a dog on the street''
(see more details and illustrative images in~\cref{sec:instructp2p}). 
%
In our work, we 
build on
Stable Diffusion~\cite{rombach2022high}, demonstrating its ability to realistically insert objects into an image without requiring further training  \textit{when combined with other components} (our POC pipeline).

Other works have explored the usage of Stable Diffusion~\cite{rombach2022high} to handle OOD classes.
Similar to us, Du~\etal~\cite{du2023dream} use text-to-image models to generate OOD images for classification and Karazija~\etal~\cite{karazija2023diffusion} for zero-shot semantic segmentation. 
While the latter~\cite{karazija2023diffusion} relies on a frozen feature extractor and focuses on zero-shot segmentation, we \textit{extend} an existing dataset with new classes.
Different from the former~\cite{du2023dream}, we target OOD in segmentation and rather than generating fully OOD images, we insert OOD objects into images realistically. Concurrent to our work, Loiseau \etal \cite{loiseau2023reliability} also propose to leverage generative models to evaluate the reliability of semantic segmentation models, while their work focuses more broadly on evaluating different aspects of uncertainty estimation, we focus on anomaly segmentation and dataset extension.
\section{Placing Objects in Context (POC)}
\label{sec:poc}

\begin{figure*}[t]
    \centering
    \includegraphics[width=1\linewidth]{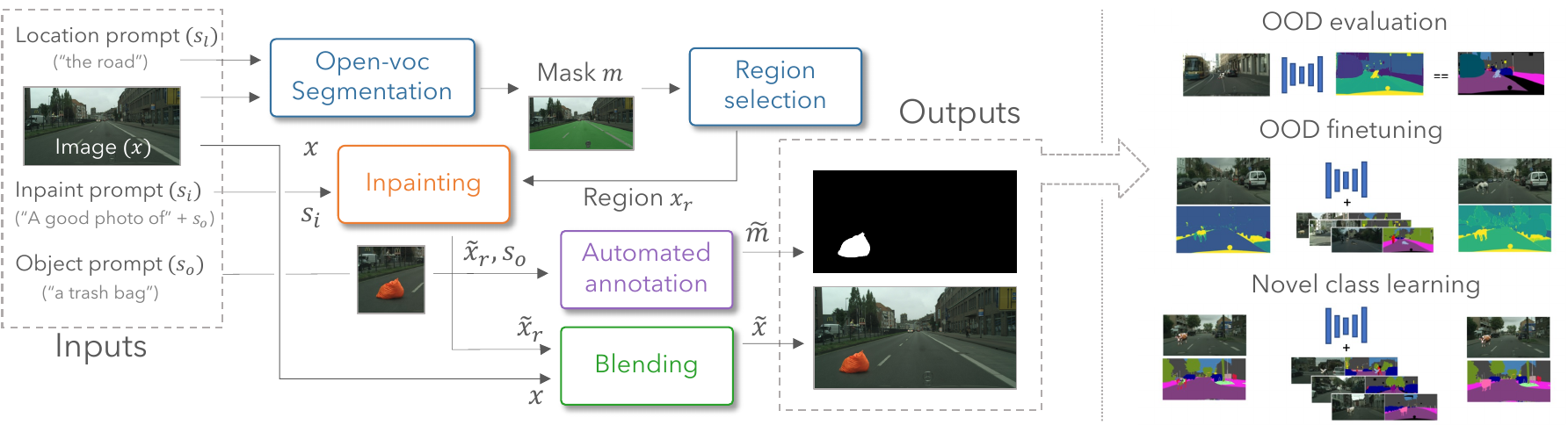}
    \caption{\small \textbf{Illustratation of our POC pipeline and applications.} Our pipeline builds on top of inpainting and open-vocabulary segmentation models to insert arbitrary objects into images realistically. The resulting images can be used for different tasks.}
    \label{fig:pipeline_fig}
\end{figure*}

We now present our proposed pipeline, Placing Objects in Context (POC). 
To recap, our desiderata to generate OOD datasets are \textit{i)} minimal shift with respect to the training set, \textit{ii)} realism, \textit{iii)} dynamic generation and \textit{iv)} low set-up costs.

\subsection{The POC pipeline}
\label{sec:poc_pipeline}

Our pipeline builds on top of two open-source models 
with permissive licenses (an important detail towards open research): 
an inpainting model from Stable Diffusion (SD~\cite{rombach2022high}) and an open-vocabulary segmentation model (GSAM) \cite{Grounded-SAM_Contributors_Grounded-Segment-Anything_2023} based on SAM \cite{kirillov2023segment} and GroundingDINO \cite{liu2023grounding}. 
In the following, we detail the main stages of our pipeline. See~\cref{fig:pipeline_fig} for a comprehensive overview.

\myparagraph{\textcolor{sb_blue}{Selecting a region for inpainting}} To realistically insert objects, it's crucial to find suitable locations for them (\ie a cat should not appear to be levitating). We use GSAM~\cite{Grounded-SAM_Contributors_Grounded-Segment-Anything_2023} to segment a suitable area for inserting the object based on a location prompt $s_l$, \eg ``the road''. 
Within this valid area, we select a region $r$ with random size and location based on user-specified limits. To assess the gain in realism obtained by guiding the object location \vs picking a random location, we conducted a human study where participants were asked to choose between image pairs with/without guided location. We find that 43\% of times the guided location was preferred \vs 18\% for the random location; 39\% of cases the preference was unclear---which is reasonable, 
since the road category tends to occupy a large percentage of the image.
Additionally, note that generated objects differ when inpainting in different locations, which makes the comparison harder.
We refer to \cref{app:ablation_location} for more details and illustrative images.

\myparagraph{\textcolor{sb_orange}{Object inpainting}}
After selecting $r$, we crop a square around it ($x_r$) and apply SD to obtain $\Tilde{x}_r$. The strong vision grounding from SD allows adding objects more realistically. For instance, we observe that the inpainting model tends to adjust the size of the object (\eg a bird will be much smaller than a garbage bin for the same region) although there is a tendency to fill all the inpainting area. We also observe that the illumination of the added object 
is adapted to the image.
This is particularly noticeable in night images from ACDC (see \cref{fig:splash}).

\myparagraph{\textcolor{sb_violet}{Automated annotation}}
In most downstream applications, we need the corresponding mask of the added object 
for training or evaluation. Again, we rely on 
open-vocabulary segmentation
to obtain $\Tilde{m}$ based on an object prompt $s_o$. 
We observed that applying GSAM to the full image often leads to false positives; we obtain better results by applying it only to the inpainted region. Note that, by relying on open-vocabulary segmentation, we can provide more accurate labels than simpler methods like foreground/background segmentation~\cite{karazija2023diffusion} that cannot distinguish between a bicycle and its rider.
During this step, we also reject images with generation failures (\eg the object was not generated or it was very unrealistic) if GSAM does not detect the generated object. 

\myparagraph{\textcolor{sb_green}{Object blending}}
While SD tends to preserve the details of the original image, it introduces slight modifications to the textures 
and sometimes noticeable changes, \eg in lane markings. To reduce undesired edits, we blend the original image $x$ and the inpainted one $\Tilde{x}$ as:
$\Tilde{x} := (1 - m) \odot x + m \odot \Tilde{x}$
where $:=$ indicates an update operation, $\odot$ is the element-wise product and $m = \mathcal{G}(\Tilde{m})$ is the object mask convolved with a Gaussian kernel. This allows for a smooth transition 
between the inpainted object and the rest of the image
while preserving some realistic local modifications, like shadows or reflections. We also considered 
applying an image2image (I2I)~\cite{isola2017image} generative model with an empty prompt after 
inpainting, but a human study showed that in 71\% of
the cases I2I blending did not improve results significantly,
in 25\% it introduced artifacts that significantly reduced realism and only in 4\% of the cases participants preferred I2I blending
(see \cref{app:blending_ablation}). In light of this, we only used the gaussian blending. \\

\subsection{Generating datasets with POC}
\label{sec:poc_datasets}

Although our pipeline can be used for multiple applications, we consider two main ones: \textit{i)} extending datasets for \textit{anomaly segmentation} and \textit{ii)} learning new classes. 
In both cases, we use Cityscapes classes as the known one; regardless of the number of added classes, 
all POC-datasets have $3\times$ the size of Cityscapes---
\ie we augment each image three times with randomly sampled prompts. Note that this does not lead to any imbalance in terms of fine-tuning steps, since all training schedules have the same total amount of iterations.

\myparagraph{Generation prompts} In all datasets we follow a similar approach. Each object class has an object prompt $s_o$, or several for diversity (\eg ``car'', ``suv'', ``van'' all belong to the Cityscapes class ``car''). Then, we build the inpainting prompt as 
$s_i = $ ``A good photo of \{$s_o$\}''.
Given that the datasets we use are all for autonomous driving applications, we use the location prompt $s_l = $ ``the road” for all objects except the class ``bird”, which has unconstrained location. We found this simple approach to yield good results without 
further
prompt engineering.

\myparagraph{Generating OOD \textit{test} sets}
To evaluate anomaly segmentation methods,
we generate three 
new test sets,
namely
\textit{CS-POC}, \textit{IDD-POC} and \textit{ACDC-POC}.
We start from the Cityscapes~\cite{cordts2016cityscapes}, IDD~\cite{varma2019idd}, and ACDC\cite{sakaridis2021acdc} 
test
sets (which have increasing domain shift \wrt Cityscapes \cite{de2023reliability}) and use the same list of OOD objects to augment the images within the 
test
set.
Our motivation is to disentangle the difficulty in detecting 
OOD objects
\vs the difficulty caused by a large distribution shift on the ID classes (while the OOD classes remain the same). Another goal is to showcase that POC can be used with different datasets seamlessly. All POC datasets contain 25 different 
OOD classes 
arbitrarily chosen to be plausible anomalies in an urban environment
(\eg wild animals, garbage bags and bins, \etc). Given our OOD objects are all \textit{synthetic}, following \cite{blum2019fishyscapes} we also add ID objects (\eg cars or persons) to ensure the model is not “simply” performing synth \vs real discrimination. See the full list in \cref{sec:poc_object_list}.

Note that, while we focus on urban segmentation datasets for consistency with previous work, POC can be applied to arbitrary datasets (see~\cref{fig:splash}, right)

\myparagraph{Generating OOD \textit{fine-tuning} sets}
Recent OOD fine-tuning methods \cite{nayal2023rba, rai2023unmasking, liu2023residual} use COCO classes not present in Cityscapes to extract OOD objects. For consistency, we generate 
\textit{POC coco} using the names of COCO classes used in previous works as prompts to inpaint them with POC (as opposed to cropping and stitching from COCO images). Moreover, we generate \textit{\mbox{POC alt.}}, an alternative fine-tuning dataset with different OOD classes (the ones used in our evaluation sets) to assess the dependence of the fine-tuning methods on the OOD objects.

\myparagraph{Extending datasets to \textit{learn new classes}}
Given a dataset $\mathcal{D}$ with a set of classes $\mathcal{K}$, we consider the task of generating an \textit{extended} dataset $\tilde{\mathcal{D}}$ with a set of classes $\tilde{\mathcal{K}} = \mathcal{K} \cup \mathcal{U}$. This dataset can be used to train models that may perform well on both $\mathcal{D}$ and on samples containing the additional classes from $\mathcal{U}$.
Following the autonomous driving use-case, we extend the Cityscapes dataset with the 6 animal classes present in the PASCAL dataset (which we use 
for evaluation). Our motivation is that wild animals cause accidents \cite{saad2019loose} and being able to segment them individually (not just as anomalies) might help preventing collisions. We call this dataset \textit{POC-A}. We further add the same classes on the CS 
test
set which we refer to as \textit{CS extended}. 
Additionally, we generate POC-CS+A, where we inpaint both animal classes and Cityscapes classes.

\begin{table*}[ht]
\centering
{
\setlength{\tabcolsep}{0.5pt}
\scriptsize{
\begin{tabular}{ll|ccc|ccc|ccc|ccc}
\toprule
\multicolumn{1}{l}{} &
\multicolumn{1}{l|}{\textbf{OOD}} &
\multicolumn{3}{c|}{\textbf{FS Static}} &
\multicolumn{3}{c|}{\textbf{FS Lost\&Found}} &
\multicolumn{3}{c|}{\textbf{SMIYC Anomaly}} &
\multicolumn{3}{c}{\textbf{SMIYC Obstacle}} \\
\multicolumn{1}{l}{\multirow{-2}{*}{\textbf{FT}}} &
\multicolumn{1}{l|}{\textbf{data}} &
F1\tiny{\tiny{$\uparrow$}} &
AuPRC\tiny{$\uparrow$} &
FPR\tiny{$\downarrow$} &
F1\tiny{$\uparrow$} &
AuPRC\tiny{$\uparrow$} &
FPR\tiny{$\downarrow$} &
F1\tiny{$\uparrow$} &
AuPRC\tiny{$\uparrow$} &
FPR\tiny{$\downarrow$} &
F1\tiny{$\uparrow$} &
AuPRC\tiny{$\uparrow$} &
FPR\tiny{$\downarrow$} \\
\midrule
&
No ft. &
65.4 &
60.1 &
7.4 &
45.5 &
36.8 &
13.9 &
88.6 &
93.0 &
3.9 &
69.3 &
73.6 &
6.9 \\
&
COCO &
\textbf{82.7} &
\textbf{88.4} &
\underline{2.2} &
70.9 &
64.5 &
13.1 &
\underline{90.5} &
{\cellcolor{gray!25} \textbf{94.7}} &
\underline{3.3} &
90.0 &
94.8 &
0.4 \\
&
POC alt. &
\underline{82.6} &
\underline{87.4} &
3.1 &
\underline{74.5} &
\underline{68.8} &
\underline{11.4} &
{\cellcolor{gray!25} \textbf{92.2}} &
\underline{93.8} &
{\cellcolor{gray!25} \textbf{2.1}} &
\underline{90.8} &
\underline{95.3} &
\underline{0.3} \\
\multirow{-4}{0.8cm}{\textbf{M2A \\[.4\baselineskip] \cite{rai2023unmasking}}} &
POC c. &
82.0 &
87.0 &
\textbf{2.1} &
{\cellcolor{gray!25} \textbf{76.5}} &
\textbf{73.0} &
\textbf{9.2} &
88.8 &
92.1 &
8.4 &
\textbf{91.4} &
\textbf{96.0} &
\textbf{0.1} \\
\midrule
&
No ft. &
21.0 &
13.9 &
38.5 &
5.4 &
1.6 &
66.3 &
50.2 &
53.0 &
39.6 &
46.0 &
45.2 &
3.2 \\
&
COCO &
83.8 &
89.6 &
1.2 &
58.0 &
58.4 &
3.2 &
\textbf{73.4} &
\textbf{78.3} &
\textbf{18.1} &
\underline{89.7} &
\textbf{94.3} &
\textbf{0.3} \\
&
POC alt. &
\underline{88.4} &
\underline{94.4} &
\underline{0.7} &
\textbf{70.3} &
\textbf{74.2} &
\underline{2.7} &
\underline{69.4} &
\underline{78.2} &
\underline{26.9} &
\textbf{89.8} &
\underline{93.6} &
\underline{0.9} \\
\multirow{-4}{0.8cm}{\textbf{RPL \\[.4\baselineskip]\cite{liu2023residual}}} &
POC c. &
{\cellcolor{gray!25} \textbf{88.5}} &
{\cellcolor{gray!25} \textbf{95.1}} &
{\cellcolor{gray!25} \textbf{0.5}} &
\underline{69.5} &
\underline{69.2} &
{\cellcolor{gray!25} \textbf{1.6}} &
62.7 &
68.0 &
46.8 &
87.8 &
92.3 &
\underline{0.9} \\
\midrule
&
No ft. &
58.2 &
59.2 &
17.7 &
63.7 &
61.0 &
10.6 &
\textbf{86.5} &
86.9 &
86.4 &
92.8 &
95.9 &
0.2 \\
&
COCO &
67.0 &
72.2 &
4.0 &
69.7 &
70.9 &
8.7 &
84.7 &
\textbf{91.1} &
5.4 &
{\cellcolor{gray!25} \textbf{94.9}} &
98.2 &
{\cellcolor{gray!25} \textbf{.04}} \\
&
POC alt. &
\underline{68.3} &
\underline{77.2} &
\underline{3.4} &
\underline{74.0} &
\underline{76.7} &
\underline{5.4} &
\textbf{86.5} &
\underline{90.5} &
\textbf{4.4} &
{\cellcolor{gray!25} \textbf{94.9}} &
{\cellcolor{gray!25} \textbf{98.4}} &
{\cellcolor{gray!25} \textbf{.04}} \\
\multirow{-4}{0.8cm}{\textbf{RbA \\[.4\baselineskip]\cite{nayal2023rba}}} &
POC c. &
\textbf{68.6} &
\textbf{77.6} &
\textbf{3.3} &
\textbf{76.5} &
{\cellcolor{gray!25} \textbf{78.9}} &
\textbf{3.3} &
85.2 &
89.4 &
\underline{5.0} &
94.6 &
{\cellcolor{gray!25} \textbf{98.4}} &
{\cellcolor{gray!25} \textbf{.04}} \\

\bottomrule

\toprule
\multicolumn{1}{l}{} &
\multicolumn{1}{l|}{\textbf{OOD}} &
\multicolumn{3}{c|}{\textbf{RoadAnomaly}} &
\multicolumn{3}{c|}{\textbf{Cityscapes--POC}} &
\multicolumn{3}{c|}{\textbf{IDD--POC}} &
\multicolumn{3}{c}{\textbf{ACDC--POC}} \\
\multicolumn{1}{l}{\multirow{-2}{*}{\textbf{FT}}} &
\multicolumn{1}{l|}{\textbf{data}} &
F1\tiny{$\uparrow$} &
AuPRC\tiny{$\uparrow$} &
FPR\tiny{$\downarrow$} &
F1\tiny{$\uparrow$} &
AuPRC\tiny{$\uparrow$} &
FPR\tiny{$\downarrow$} &
F1\tiny{$\uparrow$} &
AuPRC\tiny{$\uparrow$} &
FPR\tiny{$\downarrow$} &
F1\tiny{$\uparrow$} &
AuPRC\tiny{$\uparrow$} &
FPR\tiny{$\downarrow$} \\ \midrule
&
No ft. &
54.8 &
55.7 &
52.2 &
40.9 &
35.4 &
13.1 &
45.6 &
42.9 &
12.0 &
18.6 &
10.3 &
26.7 \\
&
COCO &
77.2 &
\underline{78.9} &
\textbf{18.5} &
88.3 &
93.9 &
0.5 &
81.0 &
85.5 &
\textbf{1.0} &
70.8 &
\underline{72.8} &
\textbf{6.2} \\ 
&
POC alt. &
{\cellcolor{gray!25} \textbf{81.5}} &
\textbf{82.3} &
36.7 &
{\cellcolor{gray!25} \textbf{91.4}} &
{\cellcolor{gray!25} \textbf{95.8}} &
\textbf{0.4} &
\underline{82.8} &
\underline{87.7} &
\underline{1.1} &
\textbf{74.0} &
\textbf{74.5} &
\underline{7.6} \\
\multirow{-4}{0.8cm}{\textbf{M2A \\[.4\baselineskip] \cite{rai2023unmasking}}} &
POC c. &
\underline{78.9} &
78.0 &
\underline{24.6} &
\underline{89.1} &
\underline{94.7} &
\textbf{0.4} &
\textbf{83.1} &
{\cellcolor{gray!25} \textbf{89.1}} &
1.3 &
\underline{72.8} &
72.0 &
8.4 \\
\midrule
&
No ft. &
24.4 &
15.0 &
70.4 &
22.7 &
13.8 &
44.9 &
8.0 &
3.5 &
61.4 &
3.3 &
1.3 &
79.0 \\
&
COCO &
60.1 &
59.2 &
27.1 &
82.1 &
88.4 &
0.7 &
72.4 &
78.7 &
1.8 &
63.6 &
65.4 &
2.6 \\ 
&
POC alt. &
\textbf{66.4} &
\textbf{69.4} &
\underline{21.7} &
\underline{86.1} &
\underline{93.2} &
\underline{0.5} &
\underline{79.7} &
\underline{85.8} &
\underline{1.0} &
{\cellcolor{gray!25} \textbf{79.4}} &
{\cellcolor{gray!25} \textbf{85.7}} &
{\cellcolor{gray!25} \textbf{0.8}} \\
\multirow{-4}{0.8cm}{\textbf{RPL \\[.4\baselineskip]\cite{liu2023residual}}} &
POC c. &
\underline{61.4} &
\underline{64.2} &
\textbf{21.4} &
\textbf{87.6} &
\textbf{94.3} &
\textbf{0.4} &
\textbf{82.8} &
\textbf{90.0} &
\textbf{0.7} &
\underline{76.0} &
\underline{81.2} &
\underline{1.3} \\
\midrule
&
No ft. &
72.8 &
78.4 &
11.8 &
73.5 &
77.9 &
3.7 &
65.5 &
65.1 &
78.9 &
24.2 &
18.7 &
90.0 \\
&
COCO &
\underline{78.5} &
\underline{83.4} &
{\cellcolor{gray!25} \textbf{8.3}} &
87.2 &
92.9 &
0.5 &
79.2 &
85.3 &
1.2 &
33.4 &
32.0 &
11.2 \\ 
&
POC alt. &
\textbf{78.3} &
{\cellcolor{gray!25} \textbf{84.1}} &
{\cellcolor{gray!25} \textbf{8.3}} &
\underline{89.3} &
\underline{95.0} &
\underline{0.4} &
\underline{83.7} &
{\cellcolor{gray!25} \textbf{89.1}} &
{\cellcolor{gray!25} \textbf{0.7}} &
\underline{55.0} &
\underline{58.4} &
{\color[HTML]{111111} \textbf{8.4}} \\
\multirow{-4}{0.8cm}{\textbf{RbA \\[.4\baselineskip]\cite{nayal2023rba}}} &
POC c. &
77.3 &
83.0 &
8.8 &
\textbf{90.5} &
{\cellcolor{gray!25} \textbf{95.8}} &
{\cellcolor{gray!25} \textbf{0.3}} &
{\cellcolor{gray!25} \textbf{83.8}} &
{\cellcolor{gray!25} \textbf{89.1}} &
\underline{0.9} &
\textbf{57.6} &
\textbf{61.1} &
\underline{9.1} \\
\bottomrule

\end{tabular}
}
}

\caption{\small \textbf{Anomaly segmentation 
results
after OOD finetuning.} We use three recent OOD fine-tuning methods (FT) and report results prior to fine-tuning (\textit{No ft.}), after fine-tuning with \textit{COCO} objects and using our POC pipeline to inpaint \textit{COCO} objects (\textit{POC c.}) or an alternative set of 25 objects likely to be found on the street (\textit{POC alt.}). Best and second best numbers for each method are highlighted in \textbf{bold} and \underline{underlined}, respectively. The best number over all methods is shaded in \colorbox{gray!40}{\textbf{gray}}. Our pipeline improves performance in most cases. Moreover, \textit{COCO} fine-tuning leads to significant improvements in our POC eval sets, consistent with previous benchmarks.}
\label{tab:anomaly_scores}
\vspace{-15pt}
\end{table*}

\section{POC for anomaly segmentation}
\label{sec:results}

As already discussed, to reliably evaluate the risk of deploying a model in a certain scenario, 
we need to test on images with minimal distribution shift \wrt the original distribution and with realistic OOD objects. Similarly, we hypothesize that fine-tuning anomaly segmentation models with more realistic anomalies 
will
improve results. To test this hypothesis, we take three recent methods for OOD fine-tuning that
rely
on 
stitching
COCO objects into CS images; instead of 
stitching, 
we use
our POC pipeline to generate the 
anomalies for fine-tuning.

\subsection{Experimental setting}
\label{sec:ood_ft}

\myparagraph{Anomaly segmentation methods} \textit{RPL}~\cite{liu2023residual} learns a module to detect anomalies via contrastive learning,
on top of 
a
segmentation network that is kept frozen. This allows
improving OOD detection with minimal degradation to the closed-set performance. \textit{Mask2Anomaly} (M2A) \cite{rai2023unmasking} and \textit{RbA}~\cite{nayal2023rba} are both based on the novel Mask2Former architecture \cite{cheng2021mask2former} which performs segmentation at the mask level, \ie by grouping pixels into ``masks'' and classifying the whole masks into the closed-set categories. This significantly reduces the pixel-level noise on the anomaly scores. RbA~\cite{nayal2023rba} performs OOD fine-tuning with a squared hinge loss while M2A~\cite{rai2023unmasking} also uses contrastive learning. We use the original code with default settings for each method and only modify the fine-tuning dataset.

\myparagraph{OOD fine-tuning data} 
For each method, we consider different ways of generating the anomaly fine-tuning datasets. First, we consider a baseline where no fine-tuning occurs (\textit{No ft.} in our tables).
Then, we consider fine-tuning on \textit{COCO} stitching \cite{chan2021entropy}. When 
we fine-tune
using our POC-generate images, we consider two cases: \textit{POC coco} 
(inpainting
the same classes as \textit{COCO} stitching) and \textit{POC alt.} 
(inpainting
alternative classes, more likely to be found on the street). 

\myparagraph{Anomaly datasets and metrics}
For evaluation, we employ five commonly used datasets, namely Fishyscapes \cite{blum2019fishyscapes}, Segment Me If You Can (object and anomaly) \cite{chan2021segmentmeifyoucan}, Road Anomaly \cite{lis2019detecting} and Lost and Found \cite{pinggera2016lost} already discussed in \cref{sec:rel_work}. Additionally, we evaluate on our POC-generated 
test sets.
Following previous work \cite{chan2021entropy, liu2023residual}, we compute three different metrics: 
maximum F1 score over all thresholds ($F1^*$), Area under the Precision-Recall Curve (AuPRC) and False Positive Rate at $95\%$ recall (FPR).

\subsection{Results}

\myparagraph{POC improves OOD detection} 
In \cref{tab:anomaly_scores}, we show that fine-tuning with \textit{POC coco} is remarkably better than the \textit{No ft.} baseline despite only using synthetic anomalies.
Moreover, \textit{POC coco}  brings 
important
improvements over COCO fine-tuning in some settings. For instance, in FS Static (for RPL or RbA), in SMIYC Obstacle (for M2A or RbA) and in FS Lost \& Found (for all methods). In other settings, it is competitive with COCO fine-tuning---except in SMIYC Anomaly (for RPL), where we observe a 
significant drop. 
One reason may be
the strong domain shift in SMIYC Anomaly, limiting the benefits of using more realistic data. On the other hand, in FS Lost \& Found, which has the closest setting to Cityscapes and real OOD objects, we carry the largest improvements.

\begin{figure}[t]
    \centering
    \includegraphics[width=.97\linewidth]{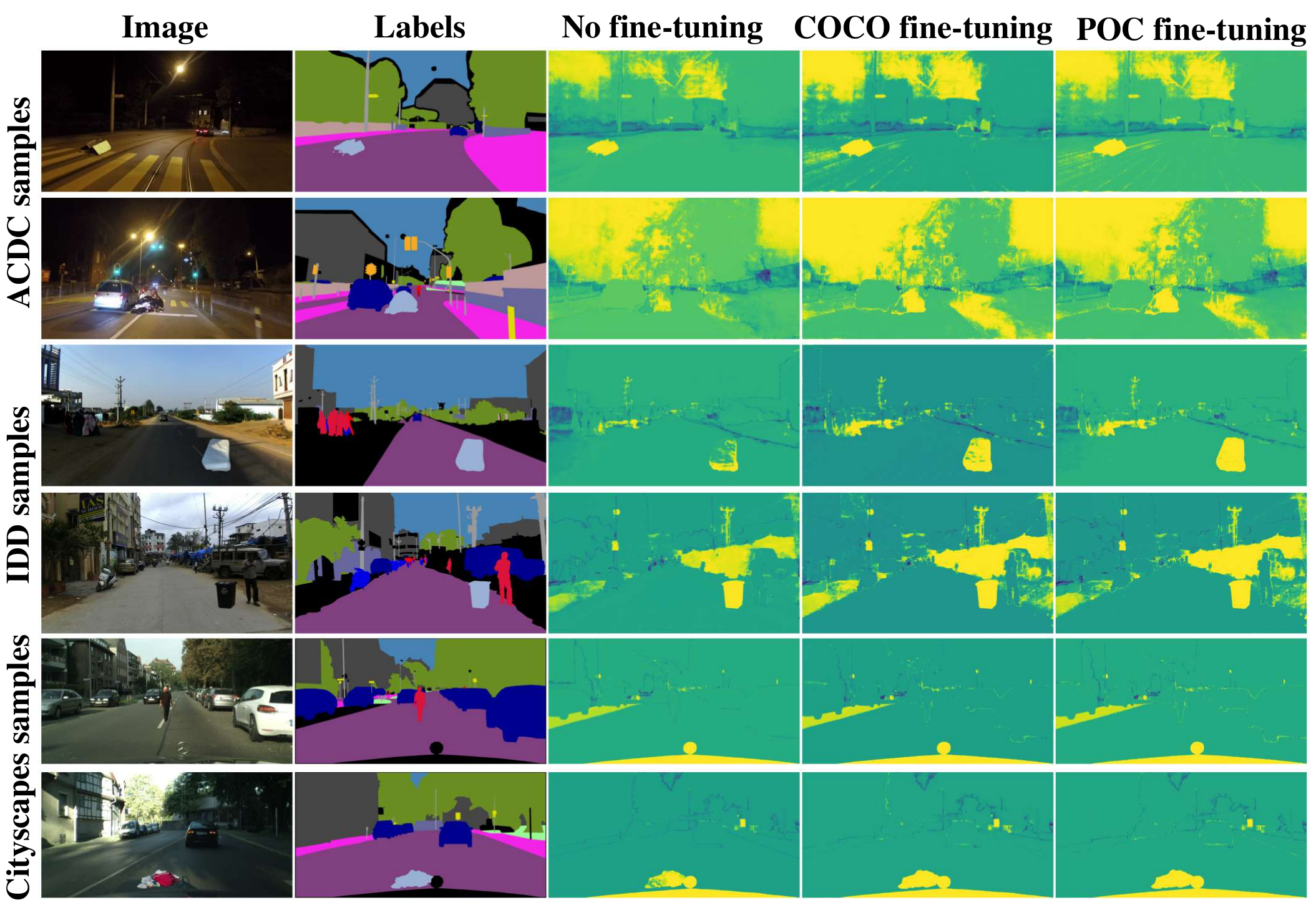}
    \caption{\small \textbf{Anomaly score maps.} Per-pixel anomaly scores on POC-generated images obtained with M2A~\cite{rai2023unmasking}, before and after fine-tuning with COCO and POC data. COCO and POC fine-tuning have notable improvements over the \textit{No ft.} baseline, \eg note the garbage bag or matress in second and third images. }
    \label{fig:confidence_maps}
\end{figure}

\myparagraph{Robustness to the choice of OOD classes} We observe that fine-tuning with \textit{POC alt.}, with different OOD classes than COCO, leads to strong results, improving over the COCO baseline in several settings and sometimes surpassing \textit{POC coco}. This shows that these methods are somewhat robust to the choice of the OOD classes. The flexibility of the POC pipeline allows studying which classes are best depending on the use-case, a possible direction for future work.
Finally, note that the best score of all methods (highlighted with gray background) corresponds to one of the POC fine-tuning in most settings.

\begin{wrapfigure}{r}{0.55\textwidth}
    \centering
    \includegraphics[width=\linewidth]{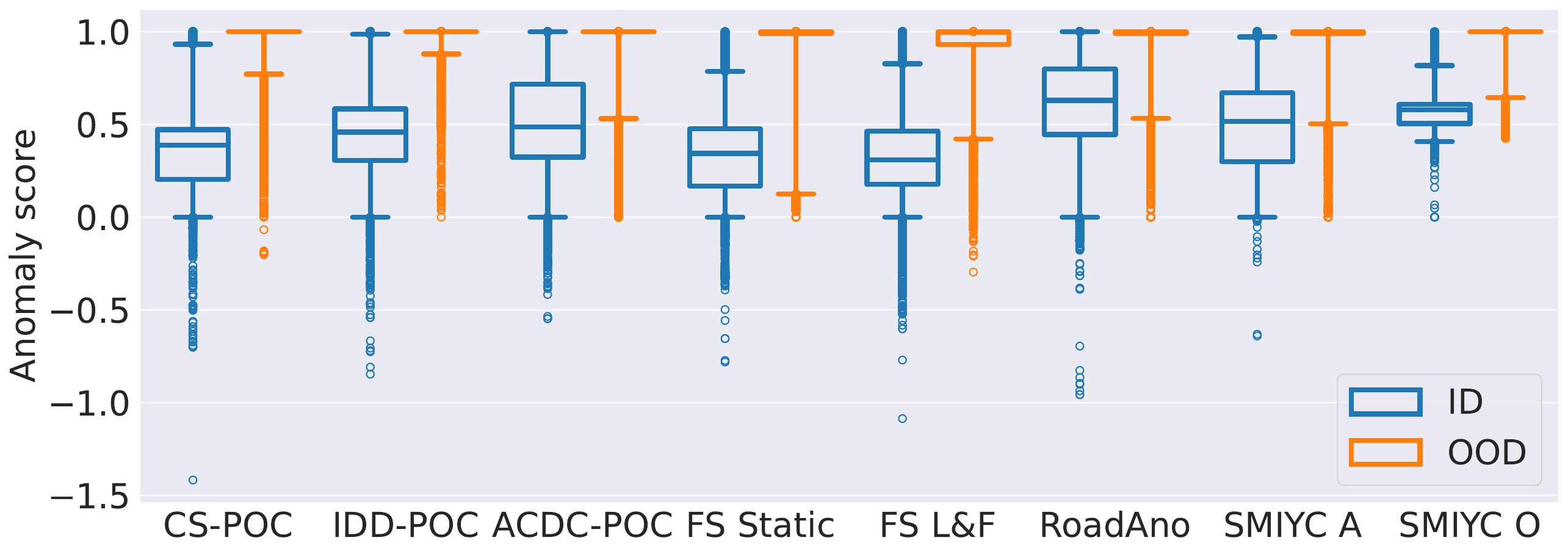}
    \caption{\small \textbf{Boxplots of anomaly scores.} All datasets have consistently very high scores for OOD pixels while ID pixels of datasets with strong distribution shifts also have shifted scores. Thus, distribution shifts may lead to underestimated performance. }
    \label{fig:histogram_confidences}
\end{wrapfigure}
\myparagraph{Performance on POC test sets}
As expected, POC fine-tuning performs best on our POC-generated test sets (see Cityscapes-POC, ACDC-POC and IDD-POC in \cref{tab:anomaly_scores}). Yet, fine-tuning with COCO also leads to notable improvements, similar to the ones observed in previous datasets. 
This suggests that POC datasets accurately reflect anomaly segmentation capabilities and can be used to efficiently build test sets.
Interestingly, we also observe that \textit{POC alt.} does not always outperform \textit{POC coco} despite sharing the same OOD classes as the POC test sets. One explanation could be that \textit{POC coco} has more OOD classes (80 compared to 25 in \textit{POC alt.}), hence, 
more diversity.

\myparagraph{Domain shifts hinder anomaly segmentation} Considering our synthetic evaluation datasets, moving from POC-CS to POC-IDD and POC-ACDC we observe a drop in performance in all methods.
Since the sets contain the same anomaly classes, the drop is due to the domain shift between 
train and evaluation
rather than hard-to-detect anomalies. Additionally, in \cref{fig:histogram_confidences} we show boxplots of the predicted anomaly scores (higher numbers reflecting 
higher chance of anomaly) for ID \vs OOD pixels. OOD pixels have very high scores independently of the dataset, while ID scores vary significantly between datasets. Datasets with strong domain shifts (\ie SMIYC Anomaly, RoadAnomaly and POC-ACDC) 
carry
larger ID anomaly scores. While anomaly segmentation under domain shift might also be an interesting task, we argue that, in practice, agents will be 
deployed in 
areas well represented by the training set.
To accurately evaluate the risk represented by anomalies, there should be no domain shift.

\begin{wrapfigure}{r}{0.6\textwidth}
    \centering
    \includegraphics[width=\linewidth]{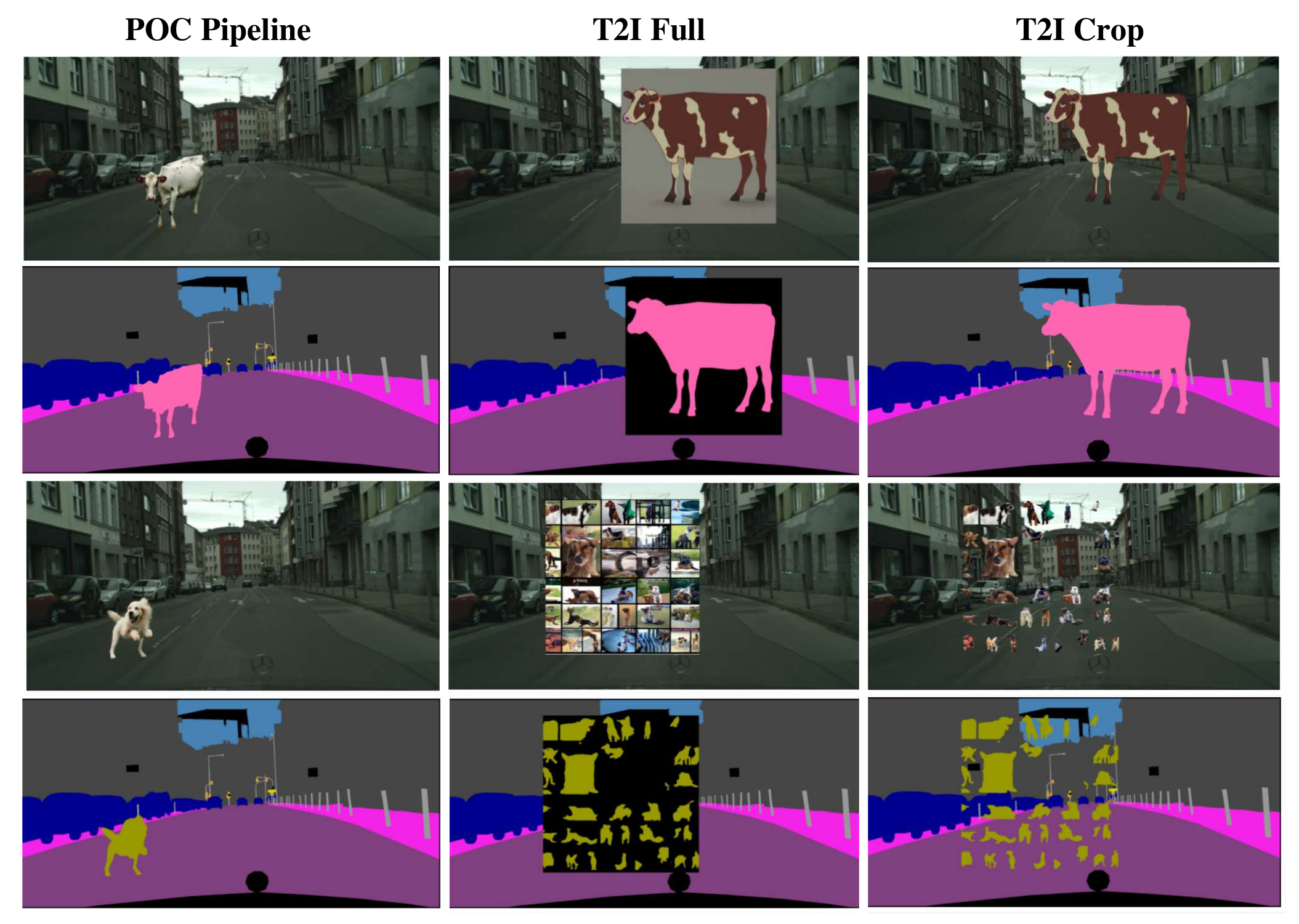}
    \caption{\small \textbf{Training image samples.} Text2Image (T2I) methods often lead to unrealistic objects. Moreover, sometimes T2I outputs are misaligned with caption, \eg ``an image of a dog'' leading to a \textit{collage} of dog images. More samples in \cref{sec:pascal_training_samples}.}
    \label{fig:pascal_training}
\end{wrapfigure}
\myparagraph{Anomaly segmentation maps} In \cref{fig:confidence_maps} we show per-pixel anomaly scores on POC-generated images computed with Mask2Anomaly~\cite{rai2023unmasking} 
prior to fine-tuning and after fine-tuning either with COCO or POC data. Aside from the inpainted anomalies (labelled in gray), we observe that in ACDC (Top) the night sky has a particularly high score and in IDD (Middle) 
a peculiar
instance of 
a 
known class (the all-road car) 
is 
also highlighted, 
potentially misleading
anomaly segmentation. More examples can be found in~\cref{sec:additional_confidence_plots}.
\section{POC to learn new classes}
\label{sec:dataset_extension}

Besides anomaly segmentation, another natural application of our POC pipeline is dataset \textit{extension}. In this final experimental section, we study this task in two dimensions: \textit{i)} adding novel objects to learn new classes and \textit{ii)} adding instances of existing classes to improve generalization.

\subsection{Experimental settings}
%
%
\myparagraph{\textit{Extended} datasets}  
Like in
our previous experiments, we use Cityscapes
as our base dataset. Motivated by autonomous driving, we are interested in learning new classes that could potentially cause road accidents if left undetected. 

\begin{wraptable}{r}{0.55\textwidth}
    \scriptsize
    \centering
    \setlength{\tabcolsep}{1pt}
    \begin{tabular}{ll|c|c|c|c}
    \toprule
{} & {} & {} & {} & \textbf{Pascal} &  \textbf{Pascal} \\
\multirow{-2}{*}{\textbf{Arch.}} & \multirow{-2}{*}{\textbf{Train set}} & \multirow{-2}{*}{\textbf{{CS}}} & \multirow{-2}{*}{\textbf{{CS ext.}}} & (A) & (CS) \\
\midrule
                           & Pascal{\tiny (b)}  & --             & --             & 80.6           & 85.8 \\
                           & CS{\tiny (b)} & 79.1           & --             & --             & 42.2 \\
                           & T2I{\tiny Full}   & 77.1           & 73.2           & 28.3           & 23.5 \\
                           & T2I{\tiny Crop}   & 78.2           & 81.5           & 26.2           & 25.1 \\
                           & POC{\tiny A}      & \textbf{80.0}  & \textbf{84.1}  & \textbf{30.4}  & 35.8 \\
\multirow{-6}{*}{\rotatebox[origin=c]{0}{DLV3+}}
& POC{\tiny CS+A}   & 79.9           & 83.8           & 28.1           & \textbf{53.6} \\ 
\midrule
                           & Pascal{\tiny (b)}     & --             & --             & 94.4           & 93.9 \\
                           & CS{\tiny (b)} & 81.6           & --             & --             & 70.5 \\
                           & T2I{\tiny Full}   & 82.3           & 83.0           & 60.4           & 65.6  \\
                           & T2I{\tiny Crop}   & 82.4           & 86.0           & 61.1           & 67.6 \\
                           & POC{\tiny A}      & \textbf{82.9}  & \textbf{86.7}  & 65.5           & 70.9 \\
\multirow{-6}{*}{\rotatebox[origin=c]{0}{CNXT}}
& POC{\tiny CS+A}   & 82.5           & 86.1           & \textbf{69.8}  & \textbf{82.4} \\
\midrule
                           & Pascal{\tiny (b)}    & --             & --             & 94.8           & 91.4 \\
                           & CS{\tiny (b)} & 76.2           & --             & --             & 79.9 \\
                           & T2I{\tiny Full}   & 77.2           & 79.5           & 82.0           & 74.3 \\
                           & T2I{\tiny Crop}   & 77.6           & 81.3           & 76.0           & 75.4 \\
                           & POC{\tiny A}      & \textbf{78.5}  & \textbf{82.3}  & 92.4           & 79.1 \\
\multirow{-6}{*}{\rotatebox[origin=c]{0}{Segm.}}
& POC{\tiny CS+A}   & 78.4           & 81.9           & \textbf{93.1}  & \textbf{89.6} \\ 
\midrule
\multirow{-1}{*}{GSAM}     & {(*)}  & 42.0         & 41.1         & 75.1          & 76.1 \\
      \bottomrule
        \end{tabular}
        \caption{\small \textbf{mIoU evaluation.} We train three 
        different models
        on Pascal and Cityscapes as baselines (b), and compare two Text2Image (T2I) generation methods with our POC. CS and A indicate Cityscapes and Pascal's animals classes, respectively. (*) We also compare with GSAM, the open vocabulary model used to automatically generate the masks of inpainted objects in POC, for completeness. GSAM is trained on
        multiple datasets \cite{Grounded-SAM_Contributors_Grounded-Segment-Anything_2023}.}
        \label{tab:pascal_mious}
\end{wraptable}

\noindent
Thus, we add \textit{animal} classes to obtain \textit{POC A} and \textit{POC CS+A} (animal and Cityscapes classes). Similar to OOD detection, finding a test set is challenging (\ie Cityscapes images with animals). 
Inspired by Karazija~\etal~\cite{karazija2023diffusion}, we use Pascal~\cite{pascal-voc-2012} animal classes to assess performance of POC-trained models. For completeness, we also evaluate on the 
test set
of Cityscapes \textit{extended} with POC \textit{(CS ext.)}.

\myparagraph{T2I baseline} Karazija~\etal~\cite{karazija2023diffusion} use text2image diffusion models to cluster the features of a frozen model for zero-shot segmentation. 
As this work closely relates to our task,
for completeness
we include two augmented datasets in our 
experiments: \textit{T2I Full} and \textit{T2I Crop} where we take the Full
(or cropped)
image generated by a T2I model and stitch it into Cityscapes images (we use the text2image model from the same work used in our POC~\cite{rombach2022high}). \cref{fig:pascal_training} shows a comparison of the synthetic datasets. T2I baselines often lead to unrealistic object size and position. Moreover, T2I models often generate images misaligned with our goal (\eg a composition of many small dog images). We generate labels with the same open-vocabulary model as our POC pipeline, which is more flexible than foreground/background segmentation---suggested in~\cite{karazija2023diffusion}.

\myparagraph{Architectures} We perform experiments using three different 
architectures: \textit{DLV3+} \cite{chen2017deeplab} with a ResNet101 backbone \cite{he2016deep}; \textit{ConvNext}\cite{liu2022convnet},
a recent convolutional architecture with UPerNet \cite{xiao2018unified}; and \textit{Segmenter}~\cite{strudel2021segmenter}, a recent transformer architecture designed for segmentation. We use the default training settings for each model. Note that our goal is not to compare architectures, but rather to evaluate the various methods employed to expand the datasets. 

\subsection{Results}

\myparagraph{POC performs better than T2I baselines} 
In \cref{tab:pascal_mious} we show the mIoU for each model after training using our different datasets, as well as two baselines trained on Cityscapes and Pascal. For all architectures, we find that training with POC datasets leads to better results than their T2I counterparts on CS extended and Pascal (A)---that is, Pascal's animal classes.
We argue that this is due to 
the more realistic
generated images 
(see
\cref{fig:pascal_training}). Interestingly, POC also improves performance on the original Cityscapes. 
Extending 
the set of classes
might act as a form of regularization,
but this requires further investigation.

\begin{figure}[t]
        \centering
        \includegraphics[width=\linewidth]{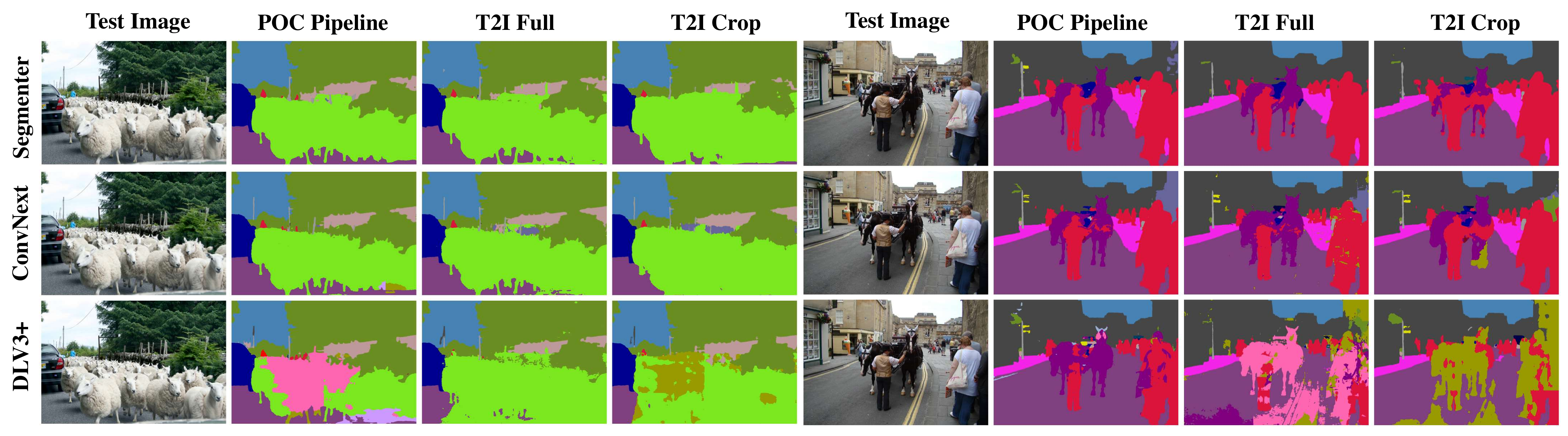}
        \caption{
        \small \textbf{Qualitative results.} We present results on web images of road scenes with animals that have a milder domain shift from Cityscapes compared to Pascal images. Qualitatively, we observe less notable differences between CNXT and Segmenter but DLV3+ is still significantly worse. More results in \cref{sec:web_images_preds}.
        }
        \label{fig:web_preds}
\end{figure}

\myparagraph{Generalization is key to learn from synthetic data} We observe that the performance of DLV3+ trained on \textit{POC A}---Cityscapes images with POC-inpainted animal classes---when evaluated on Pascal's animal classes (30.43 mIoU) is much lower than that of Segmenter (92.4 mIoU), which achieves an \textit{mIoU competitive with the baseline trained directly on Pascal} (94.75). In order to understand that gap, we evaluate the performance of these models on the Pascal classes that are present on Cityscapes (\ie car, motorcycle, bike, person, train and bus). Interestingly, we observe that when trained on Cityscapes (without any inpainted classes) DLV3+ still has a much lower mIoU than Segmenter. This shows that DLV3+ has much less generalization capability independently of the synthetic classes. One could argue that perhaps the low performance of DLV3+ on Pascal is due to the large domain shift, however, in \cref{fig:web_preds} we show qualitative results on web images of driving scenes, closer to the Cityscapes domain, and still observe that DLV3+ is notably worse than the other methods. 


We hypothesize that, in order to learn transferable features from synthetic data, the generalization capability of the model plays a key role. Indeed, although generative models have improved the realism of generated content remarkably, there is still a synth2real gap. Thus, more robust models (\ie with strong transferability of learned features) may be able to extract more useful features rather than overfitting to brittle patterns in generated data. 

We also note that the ResNet backbone in DLV3+ is pre-trained on the smaller Imagenet 1k, 
while ConvNeXt and Segmenter use Imagenet 21k; this might play a role, too. 
Nevertheless, there is also a gap between ConvNeXt and Segmenter in generalization, 
which cannot be explained by pre-training {alone}. Perhaps self-attention or image tokenization are relevant, but this would require a more in-depth analysis which is out of the scope of this work.

\myparagraph{POC to augment existing classes} When evaluating the performance of Segmenter trained on \textit{POC A} (containing the original Cityscapes classes extended with POC-added animal classes) we see a gap between Pascal (A) and Pascal (CS). This compels us to also inpaint Cityscapes classes with POC, obtaining \textit{POC CS+A}. Training with the latter dataset, leads to remarkable improvements in Pascal 
(CS classes)
in all three networks, 
significantly outperforming the baseline that was only trained on Cityscapes. Coupled with the milder but consistent improvement on the original Cityscapes, this indicates that our pipeline could also be helpful in improving in-distribution generalization.

\myparagraph{Open vocabulary baseline} 
Since we rely on a
open-vocabulary segmentation method (GSAM \cite{Grounded-SAM_Contributors_Grounded-Segment-Anything_2023}) to label objects inserted with POC, we also evaluate its performance as a baseline. Interestingly, although we find the performance
on Pascal reasonable, GSAM significantly underperforms other models on Cityscapes. We hypothesize this may be due to the one-vs-all nature of open-vocabulary predictors (where each class is predicted individually with a prompt) that does not perform well on complex images with many classes. Interestingly, 
the Segmenter model trained
with POC significantly outperforms GSAM on Pascal. 

Our
dataset extension can be interpreted as a form of knowledge distillation from a teacher model (GSAM), which has been used to label the new classes in POC-extended datasets, while significantly improving its performance.
On the other hand, note that
the test setting is very different
from the context in which
we use GSAM in the POC pipeline, since 
in the latter
we know the object that is being inpainted and only apply GSAM to the cropped region.

\section{Concluding remarks}
To accurately assess anomaly segmentation capabilities 
of models deployed
in open-world settings, we 
argue
that datasets should be realistic and 
carry
a small domain shift 
with respect to the training distribution,
as we 
show
it can hinder OOD detection. Towards  
this
goal, we 
introduce
the Placing Object in Context (POC) pipeline, 
that allows adding
\textit{any} object into \textit{any} image based on 
simple text prompting.
POC 
uses
diffusion models
and open-vocabulary segmentation to achieve high realism and versatility.

We 
showcase
POC's flexibility 
by generating
three anomaly segmentation test sets: POC-CS, POC-IDD and POC-ACDC. Moreover, we observe that generating 
data for OOD fine-tuning
with POC brings significant improvements in 
standard anomaly segmentation benchmarks.
Beyond anomaly segmentation, we use 
POC-generated 
datasets to learn new classes without any real example. Interestingly, we observe that the combination of more realistic synthetic data with recent segmentation models with strong generalization capabilities can lead to a remarkable performance, competitive with training on real data.

In future work, we hope to better understand how to select the optimal 
set of anomalies for fine-tuning and how modern architectures can effectively rely on synthetic data to learn new classes. 




\clearpage  

\section*{Acknowledgements}
This work is supported by the UKRI grant: Turing AI Fellowship EP / W002981 / 1. We would also like to thank the Royal Academy of Engineering.

%
%
\bibliographystyle{splncs04}
\bibliography{main}

\clearpage
\setcounter{page}{1}
\appendix
\section{POC datasets object list}
\label{sec:poc_object_list}
Our POC evaluation sets (\ie POC-CS, POC-IDD and POC-ACDC) are obtained by adding objects to different self-driving datasets. Following works like Lost and Found \cite{pinggera2016lost}, we compiled a list of 25 objects that 
can be found
on the road. 

The \textbf{anomaly list} is as follows:
``stroller'',
``trolley'',
``garbage bag'',
``wheelie bin'',
``suitcase'',
``skateboard'',
``chair dumped on the street'',
``sofa dumped on the street'',
``furniture dumped on the street'',
``matress dumped on the street'',
``garbage dumped on the street'',
``clothes dumped on the street'',
``cement mixer on the street'',
``cat'',
``dog'',
``bird flying'',
``horse'',
``skunk'',
``sheep'',
``crocodile'',
``alligator'',
``bear'',
``llama'',
``tiger'' and
``monkey''.

Additionally, we also add a few classes from Cityscapes to make sure that anomaly segmentation models are
indeed detecting \textit{anomalies} and not 
merely identifying \textit{synthetic objects}.

\textbf{Cityscapes classes} inluded are: ``rider'', ``bicycle'', ``motorcycle'', ``bus'', ``person'' and ``car''.

\section{Anomaly segmentation methods: mIoU on Cityscapes}
\label{sec:miou_performance}

Although fine-tuning methods with OOD samples can improve anomaly segmentation significantly, they 
may affect the closed-set
performance. In \cref{tab:mious_ood} we report the mIoU on Cityscapes of all methods reported in the main paper, 
showing that
fine-tuning with POC data 
does not negatively impact closed-set performance.
\begin{table*}[h]
\centering
\scriptsize
\setlength{\tabcolsep}{1pt}
\begin{tabular}{@{}l|cccc|cccc|cccc@{}}
\toprule
\textbf{Method}          & \multicolumn{4}{c|}{\textbf{Mask2Anomaly}} & \multicolumn{4}{c|}{\textbf{RPL}} & \multicolumn{4}{c}{\textbf{RbA}} \\ \midrule
\textbf{OOD data}        & No ft.        &  coco   & POC c    & POC alt.    & No ft.        &  coco   & POC c    & POC alt.    & No ft.        &  coco   & POC c    & POC alt.\\
\textbf{mIoU $\uparrow$} & 78.29     & 78.34    & 78.33    & 78.49    & 90.94  & 90.94  & 90.94  & 90.94  & 82.25  & 82.15  & 82.17  & 82.16 \\ \bottomrule
\end{tabular}

\caption{\small \textbf{mIoU on Cityscapes validation set.} We compute the mIoU after fine-tuning with different datasets (complementing results in \cref{tab:anomaly_scores}). We observe that fine-tuning with our POC datasets does not degrade the closed-set performance. 
}
\label{tab:mious_ood}
\end{table*}

\section{Adding objects with Instruct Pix2Pix}
\label{sec:instructp2p}

\begin{figure}[ht]
    \centering
    \includegraphics[width=0.85\linewidth]{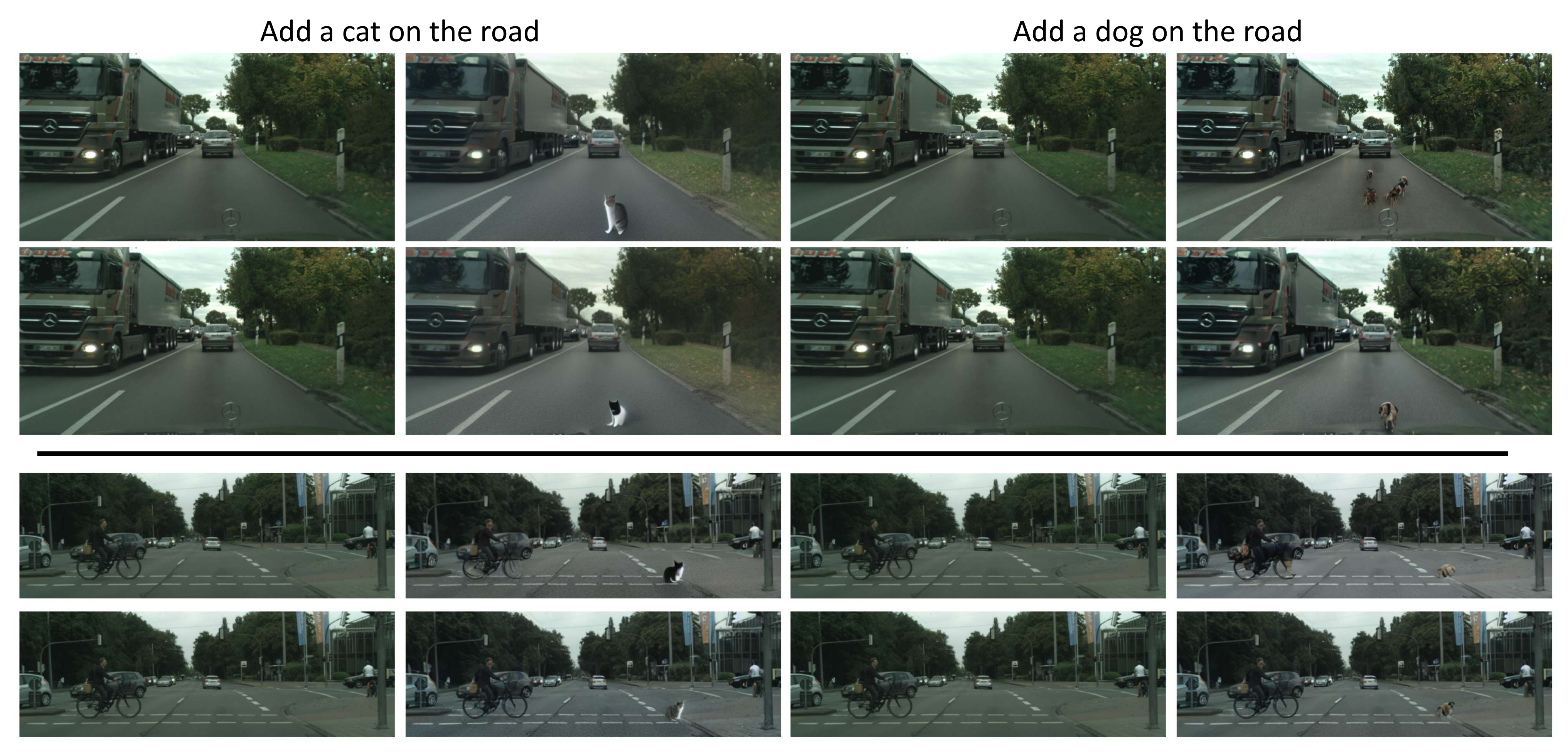}
    \caption{\textbf{Sample images from InstructPix2Pix \cite{brooks2023instructpix2pix}.} We observe that InstructPix2Pix has a bias to replace certain objects or features in the scene. In top images it replaces the mercedes logo of the ego vehicle while in bottom images it replaces the edge of the sidewalk.}
    \label{fig:ip2p}
\end{figure}

When building our pipeline, we explored different generative methods. In particular, InstructPix2Pix (IP2P) \cite{brooks2023instructpix2pix} has showed remarkable performance following natural language instructions (\eg ``turn the sofa red''). Therefore, it would be natural to have such ``general-purpose editing'' methods as baselines to add objects to the images. We observed that IP2P seems to be biased towards \textit{modifying} objects in the scene rather than \textit{adding new ones}. In \cref{fig:ip2p} we show several examples of images generated with IP2P. In our initial experiments, we observe how new objects tend to replace the 
logo of the ego vehicle (top images). If we remove the bottom of the image (middle section), we then observe that some particular image features (\eg the edge of the sidewalk) tend to be replaced. Moreover, the added objects tend to lack realism and, given that the changes are not constrained to a particular region, 
editing via IP2P usually results
in undesired modifications in other image regions.



\section{Ablation of guided region selection} \label{app:ablation_location}
We argued in the main text that in order to add objects into scenes realistically, it is important to properly place them. Thus, we apply GSAM to segment a valid area based on a location prompt (\eg ``the road'') and then select a region randomly within the valid area. Without this component in our pipeline,  the objects result inpainted in clearly unrealistic positions.

\begin{figure}[ht]
    \centering
    \includegraphics[width=1\linewidth]{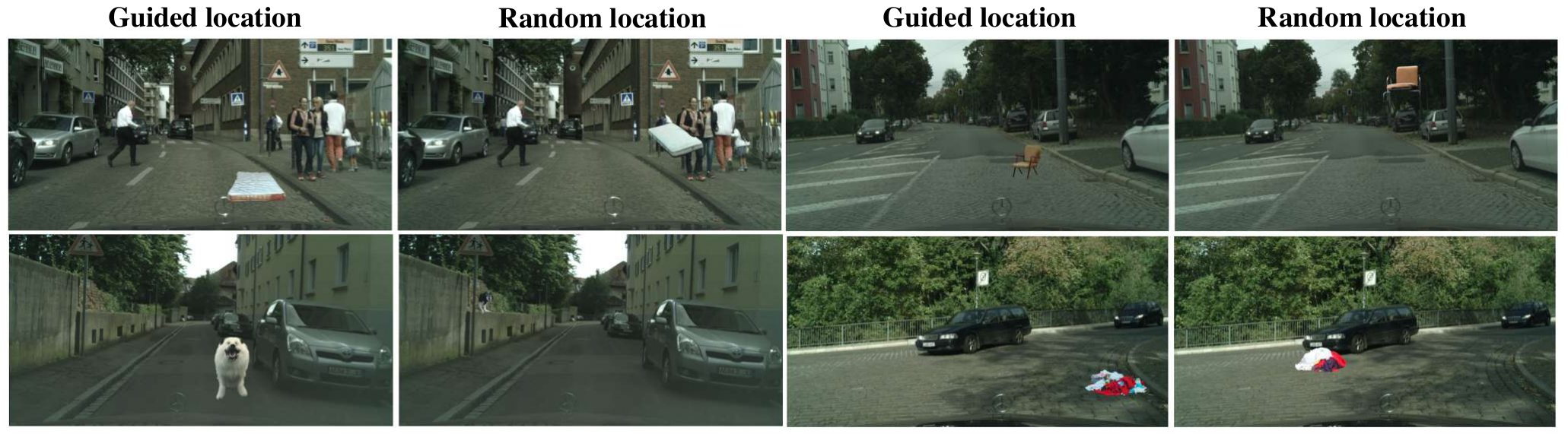}
    \caption{\textbf{Location ablation examples:} Different examples of images inpainted with our guided location or random location of objects. We observe how in many cases, random location leads to unrealistic scenes.}
    \label{fig:ablation_location}
\end{figure}

To assess the realism introduced by guiding the object location \vs placing objects randomly, we conducted a human study where participants were shown different pairs of images, one with guided location inpainting and the other with random placement, and asked to choose the most realistic image in each pair. We observed that $39\%$ of times the preference was unclear, $43\%$ guided location was preferred and $18\%$ random location was preferred. Note that a large portion of Cityscapes images is road/street, thus, a significant portion of randomly placed objects will be realistic. On the other hand, when the location is different, the generated objects also vary (even if fixing the random seed), which adds some noise to the study. All in all, we do observe a clear preference for guided location compared to random placement. In \cref{fig:ablation_location} we show some examples of image pairs with guided and random locations. Note how the added ``dumped clothes'' in the bottom right are both in realistic locations while the other objects are placed unrealistically with the random location. 

\section{Ablation of image2image blending} \label{app:blending_ablation} Similar to our location ablation, we also study if applying an image2image (I2I) model after object inpainting leads to better blending. In particular, we performed a human study where participants had to choose the most realistic image between our blending ans I2I. In $71\%$ of the cases I2I blending did not improve results significantly, in $25\%$ it introduced artifacts that significantly reduced realism and in only $4\%$ participants preferred I2I blending.

\begin{figure}[ht]
    \centering
    \includegraphics[width=1\linewidth]{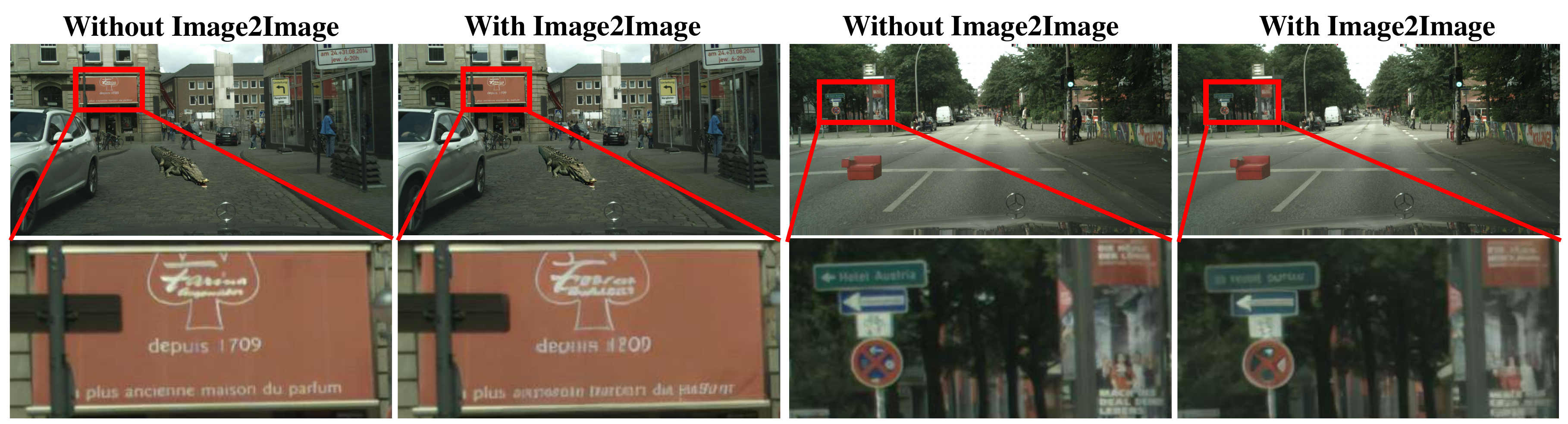}
    \caption{\textbf{Examples of object blending:} On the left image pair, we observe the presence of articles in the text of an old perfume shop which is legible on the right image but becomes illegible with I2I. On the right image pair, one can observe differences on the traffic signs. For instance, the text ``Hotel Austria'' on the green sign on the top (legible when zoomed on the left) becomes again illegible on the right image. Also, the white squared sign with a depiction of a bike without I2I becomes uninterpretable after I2I.}
    \label{fig:blending_ablation}
\end{figure}

In particular, we noted that I2I blending adds slight artefacts that can degrade the realism of the image significantly, these become especially noticeable in text or traffic signs where small variations can change the semantics drastically. In \cref{fig:blending_ablation} we present two examples of such images where artefacts are highlighted.

\clearpage
\section{AUPRC plots}
\begin{figure}[ht]
    \centering
    \includegraphics[width=1\linewidth]{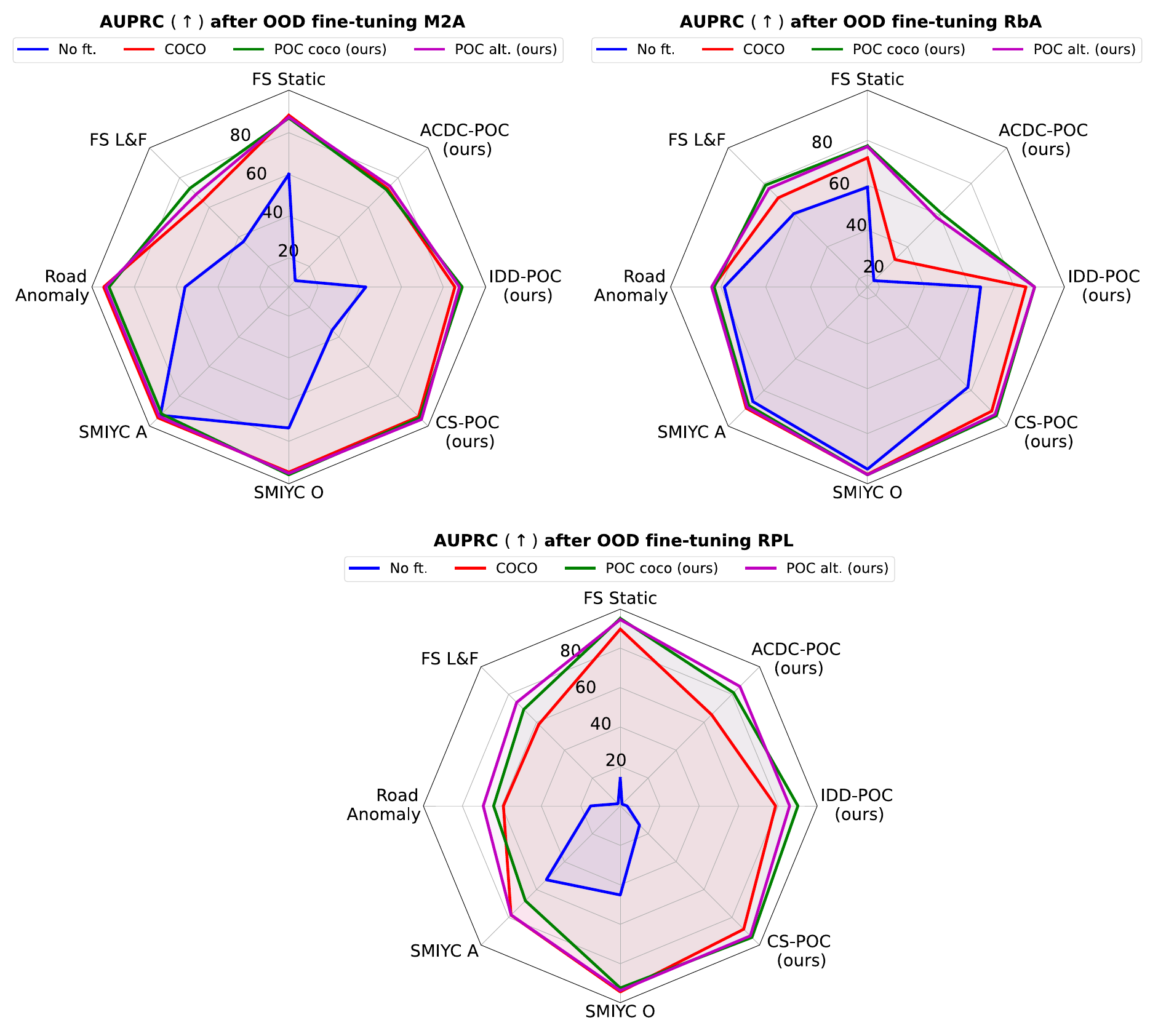}
    \caption{\textbf{AUPRC on different anomaly segmentation datasets.} We compare three different anomaly segmentation methods, M2A\cite{rai2023unmasking}, RPL\cite{liu2023residual} and RbA\cite{nayal2023rba} with different fine-tuning datasets. Fine-tuning with POC-generated images tends to bring improvements or match COCO fine-tuning in most settings.}
    \label{fig:spider_plots}
\end{figure}

In \cref{fig:spider_plots}
we visualize the AUPRC results
for all methods,
complementing the visualization in \cref{fig:splash}.

\clearpage

\section{Additional Pascal training samples}
\label{sec:pascal_training_samples}

\begin{figure}[ht]
    \centering
    \includegraphics[width=0.85\linewidth]{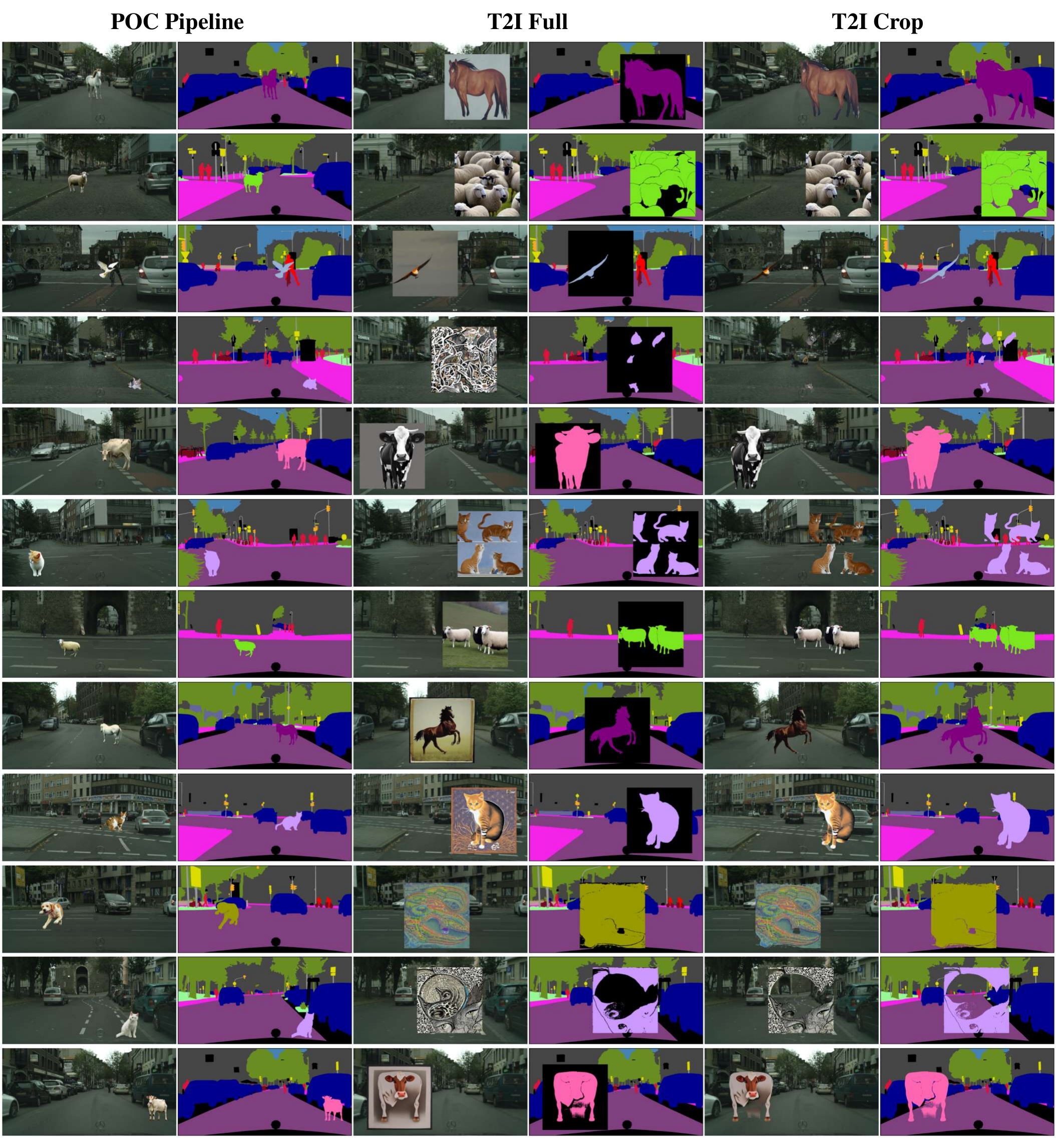}
    \caption{\textbf{Training image samples.} Additional training images to learn new classes, complementing \cref{fig:pascal_training}.}
    \label{fig:more_pascal_training}
\end{figure}

In \cref{fig:more_pascal_training} we show additional samples generated with our POC pipeline as well as T2I baselines (that use text-to-image models inspired by \cite{karazija2023diffusion}). 

\clearpage

\section{Additional anomaly score maps}
\label{sec:additional_confidence_plots}

In this appendix section we add more visualizations of anomaly score maps with different methods. We show results from our three POC-generated datasets as well as samples from previous anomaly datasets.

\begin{figure}[ht]
    \centering
    \includegraphics[width=0.85\linewidth]{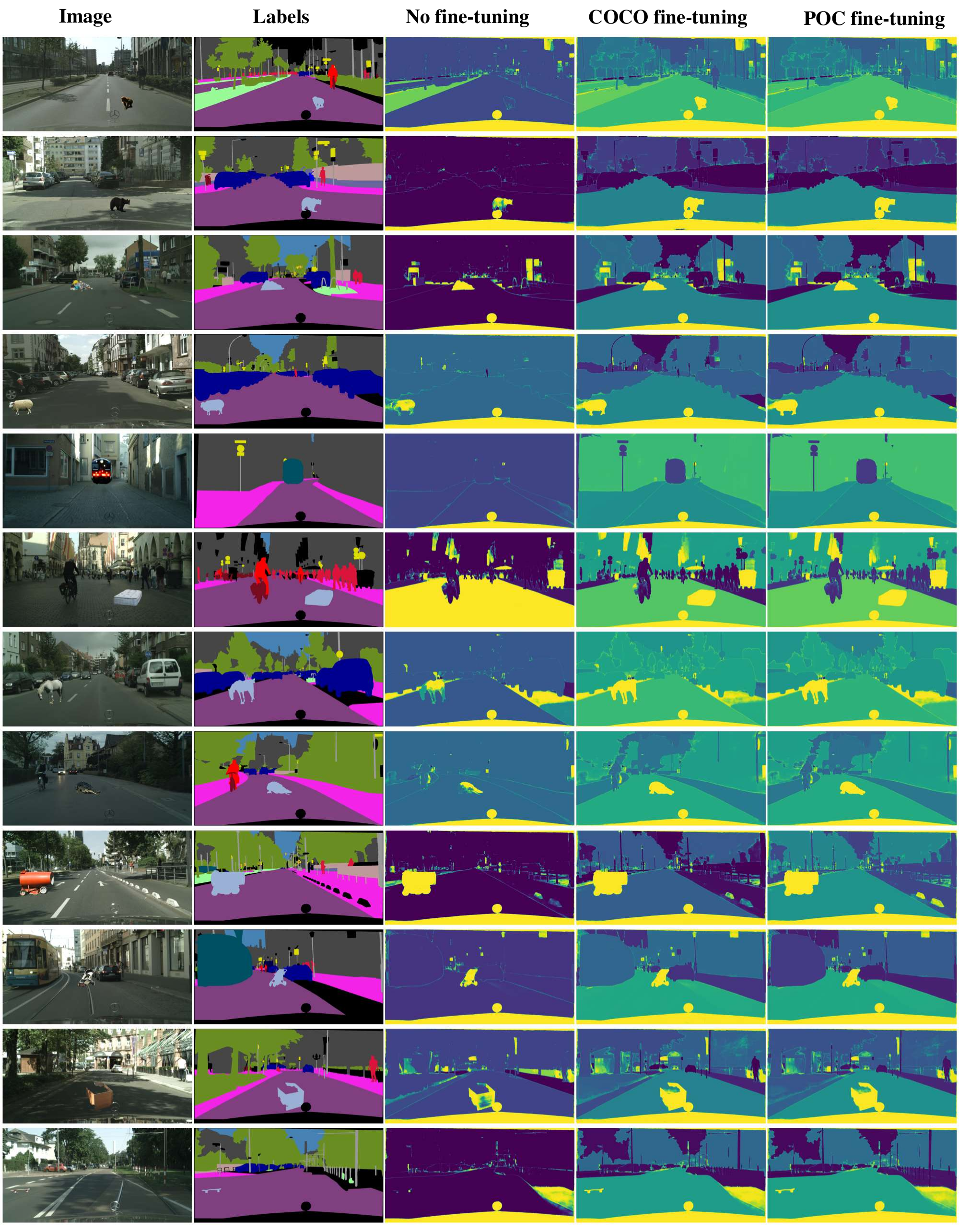}
    \caption{\textbf{M2A anomaly scores on 
    CS-POC 
    samples.}}
\end{figure} 

\clearpage

\begin{figure}[ht]
    \centering
    \includegraphics[width=0.85\linewidth]{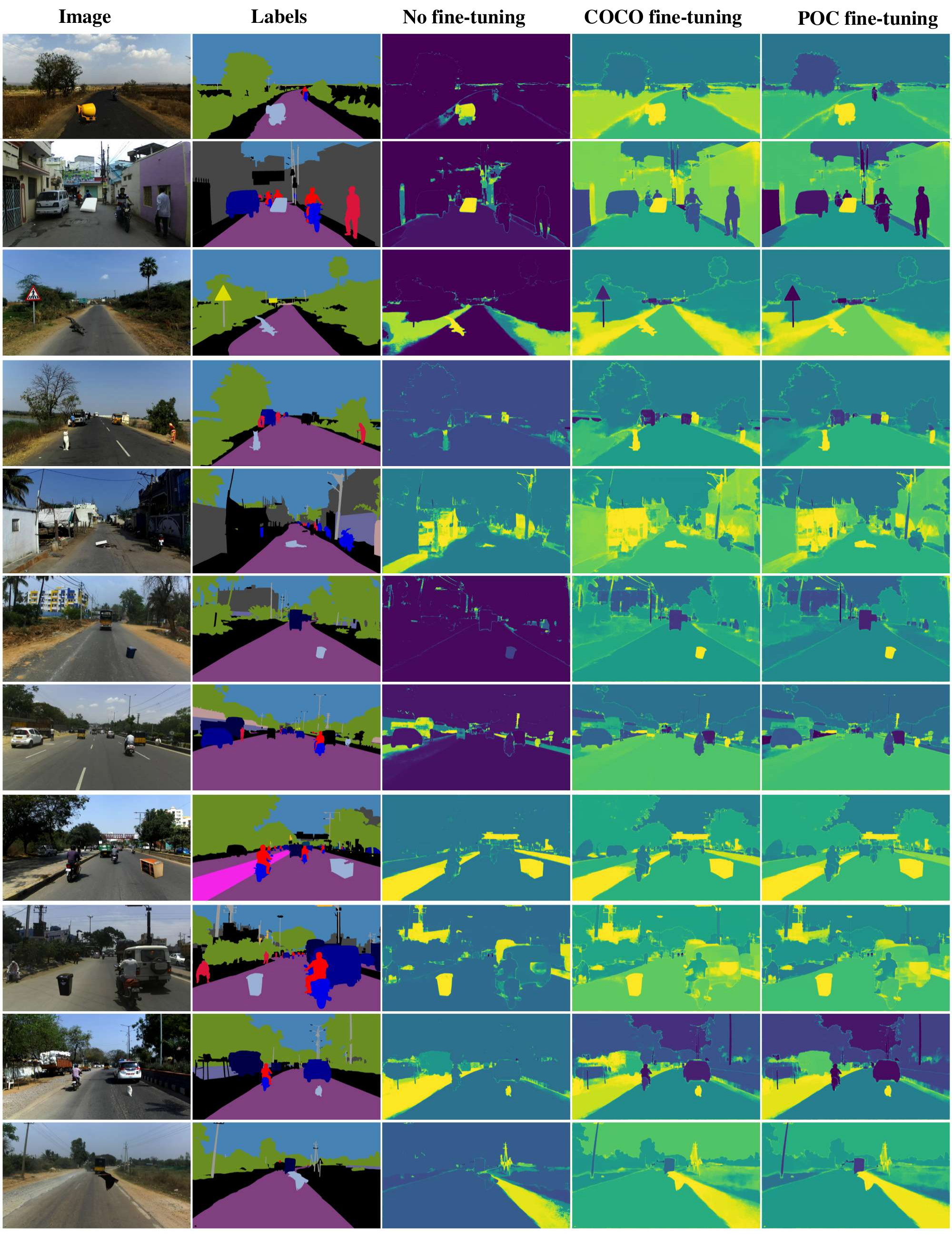}
    \caption{\textbf{M2A anomaly scores on IDD-POC 
    samples.}}
\end{figure} 

\clearpage

\begin{figure}[ht]
    \centering
    \includegraphics[width=0.85\linewidth]{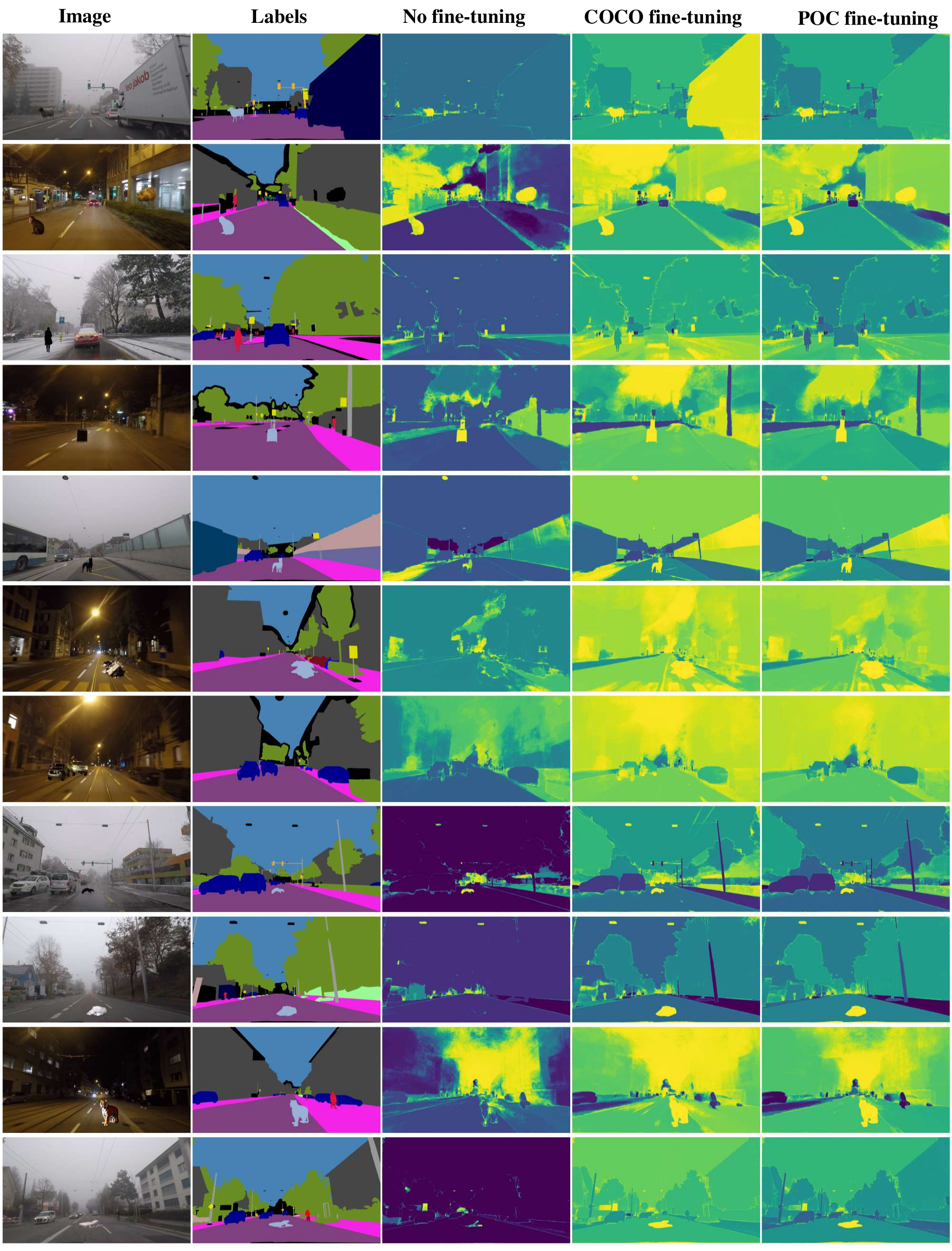}
    \caption{\textbf{M2A anomaly scores on
    ACDC-POC 
    samples.}}
\end{figure} 

\clearpage

\begin{figure}[ht]
    \centering
    \includegraphics[width=0.85\linewidth]{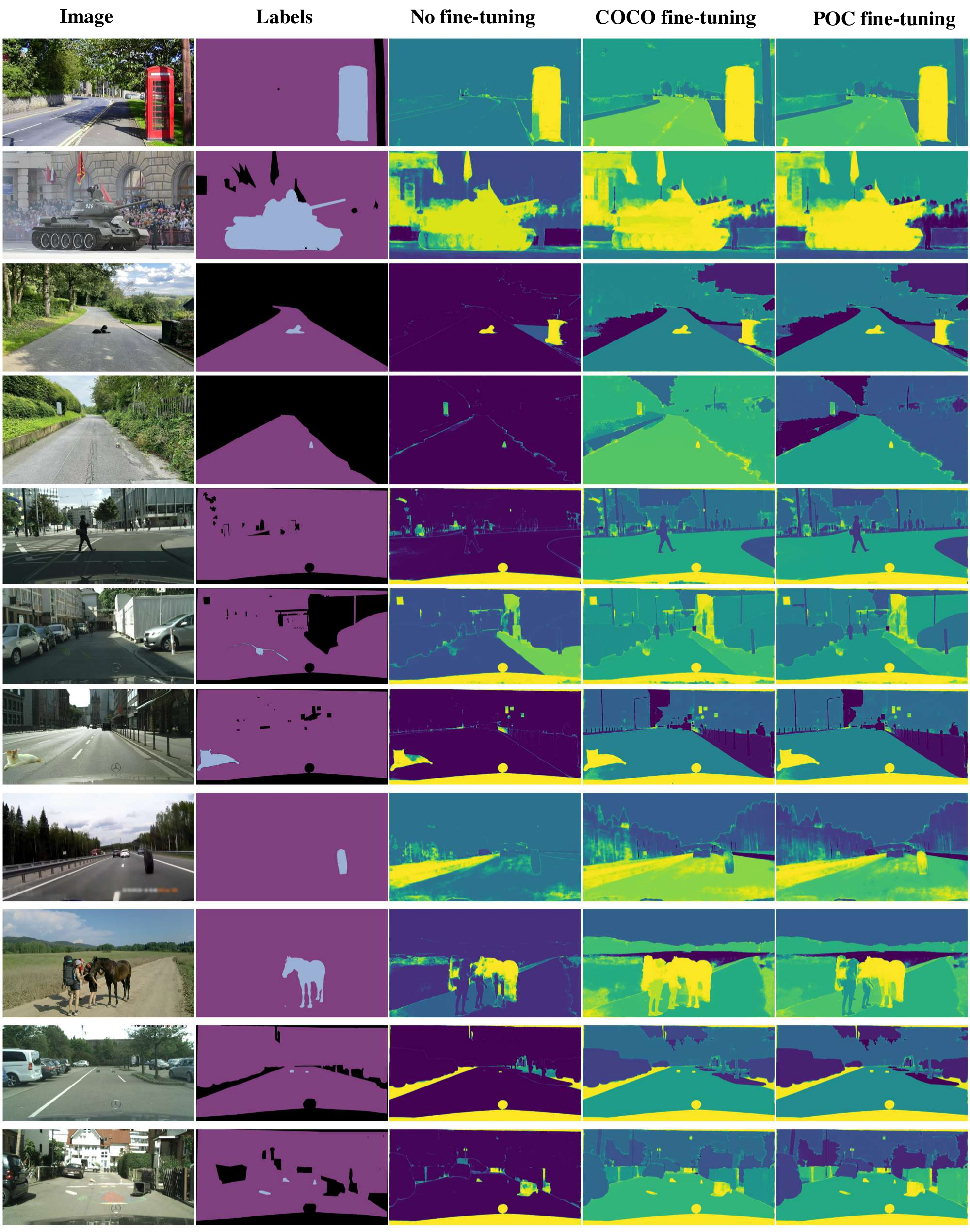}
    \caption{\textbf{M2A anomaly scores on samples from related datasets (see~\cref{fig:dset_comparison})}.}
\end{figure} 

\clearpage

\begin{figure}[ht]
    \centering
    \includegraphics[width=0.85\linewidth]{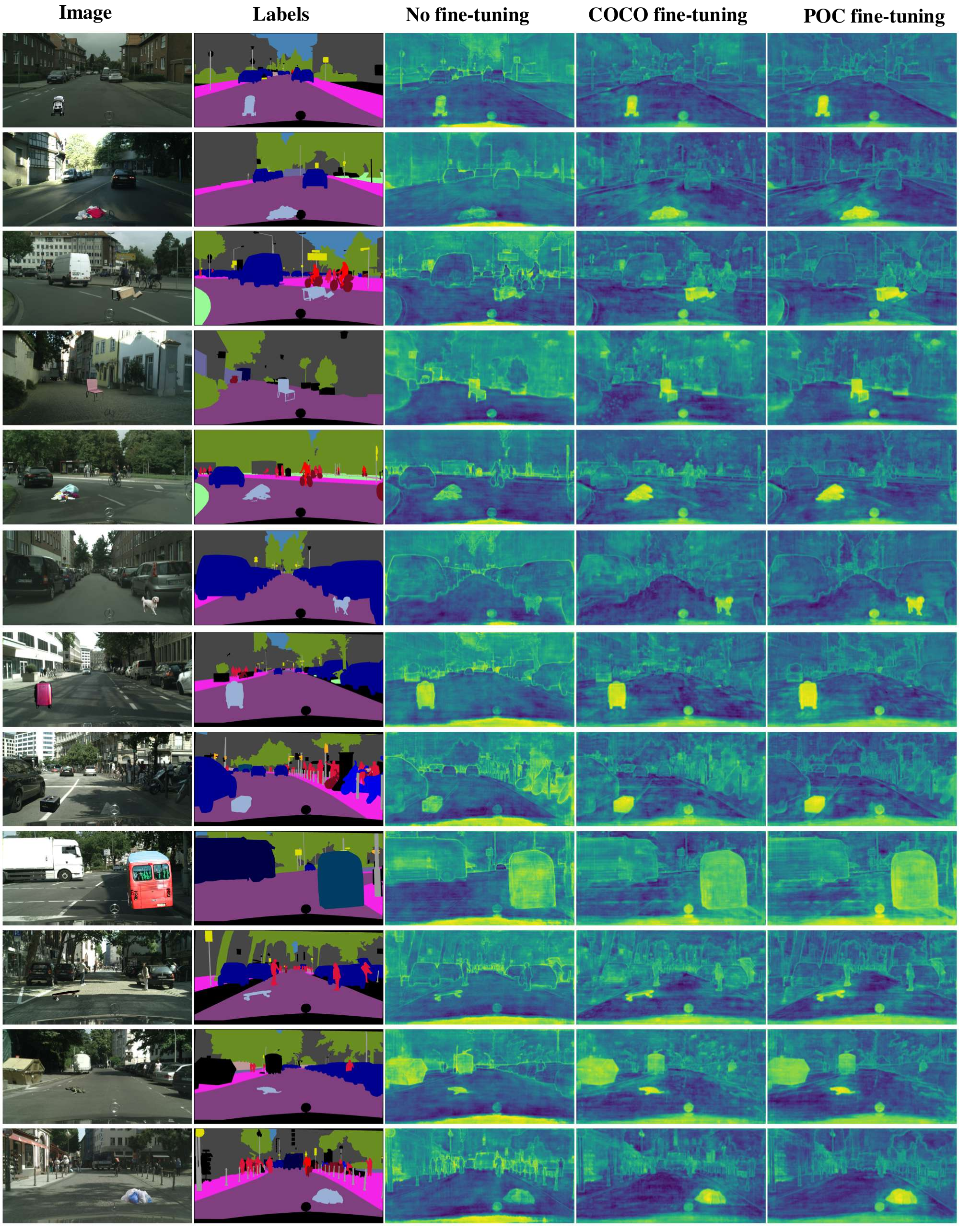}
    \caption{\textbf{RPL anomaly scores on 
    CS-POC 
    samples.}}
\end{figure} 

\clearpage

\begin{figure}[ht]
    \centering
    \includegraphics[width=0.85\linewidth]{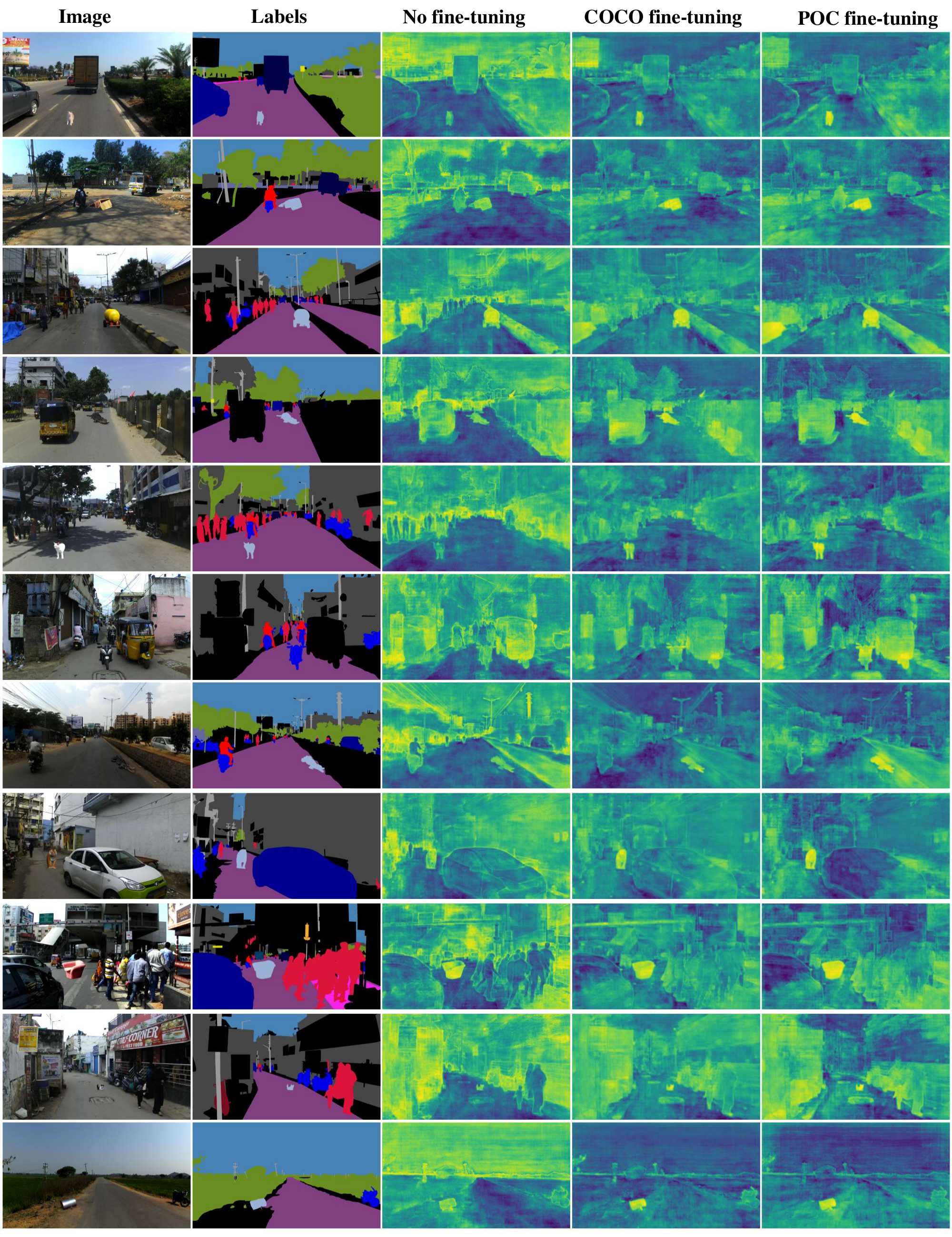}
    \caption{\textbf{RPL anomaly scores on IDD-POC 
    samples.}}
\end{figure} 

\clearpage

\begin{figure}[ht]
    \centering
    \includegraphics[width=0.85\linewidth]{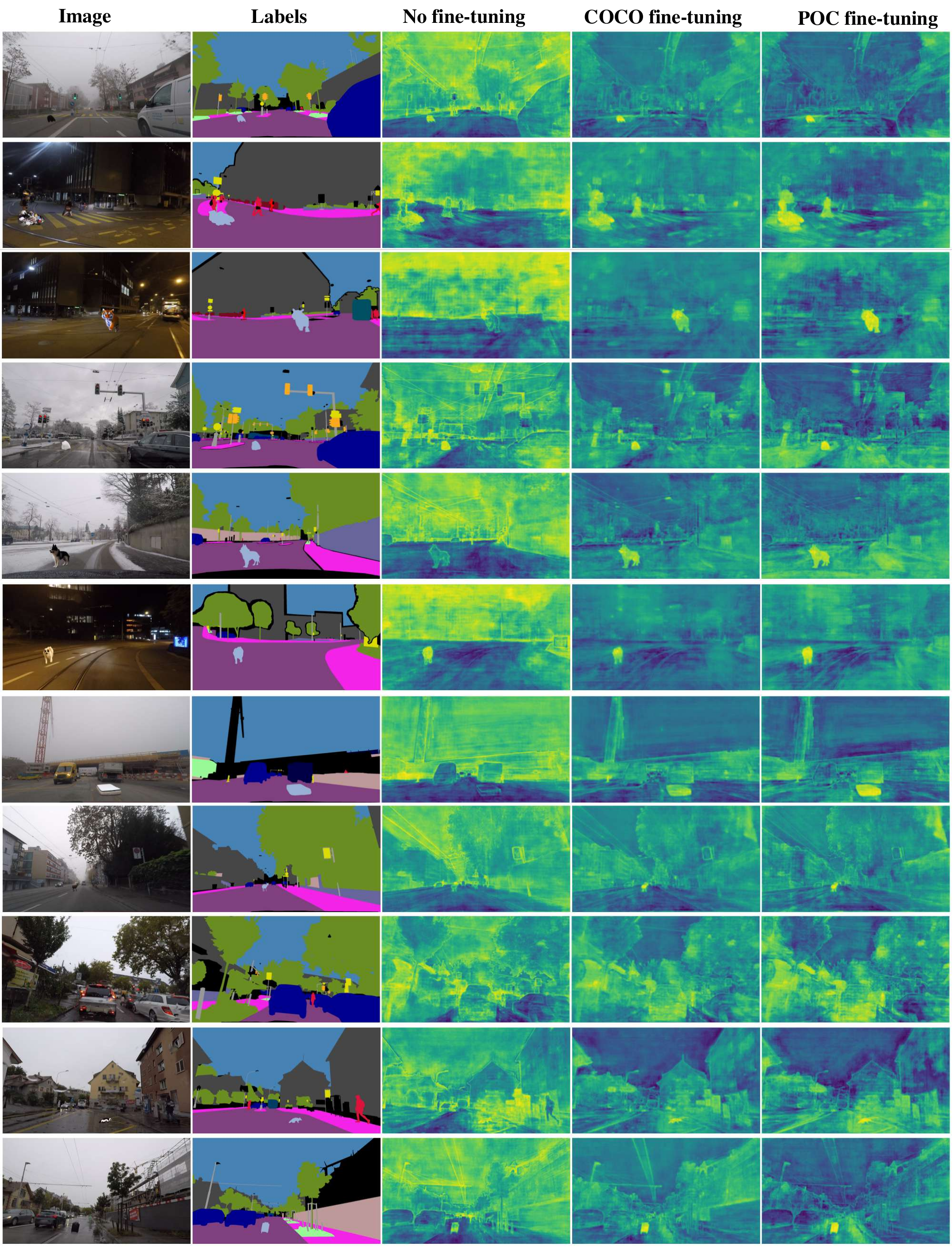}
    \caption{\textbf{RPL anomaly scores on ACDC-POC 
    samples.}}
\end{figure} 

\clearpage

\begin{figure}[ht]
    \centering
    \includegraphics[width=0.85\linewidth]{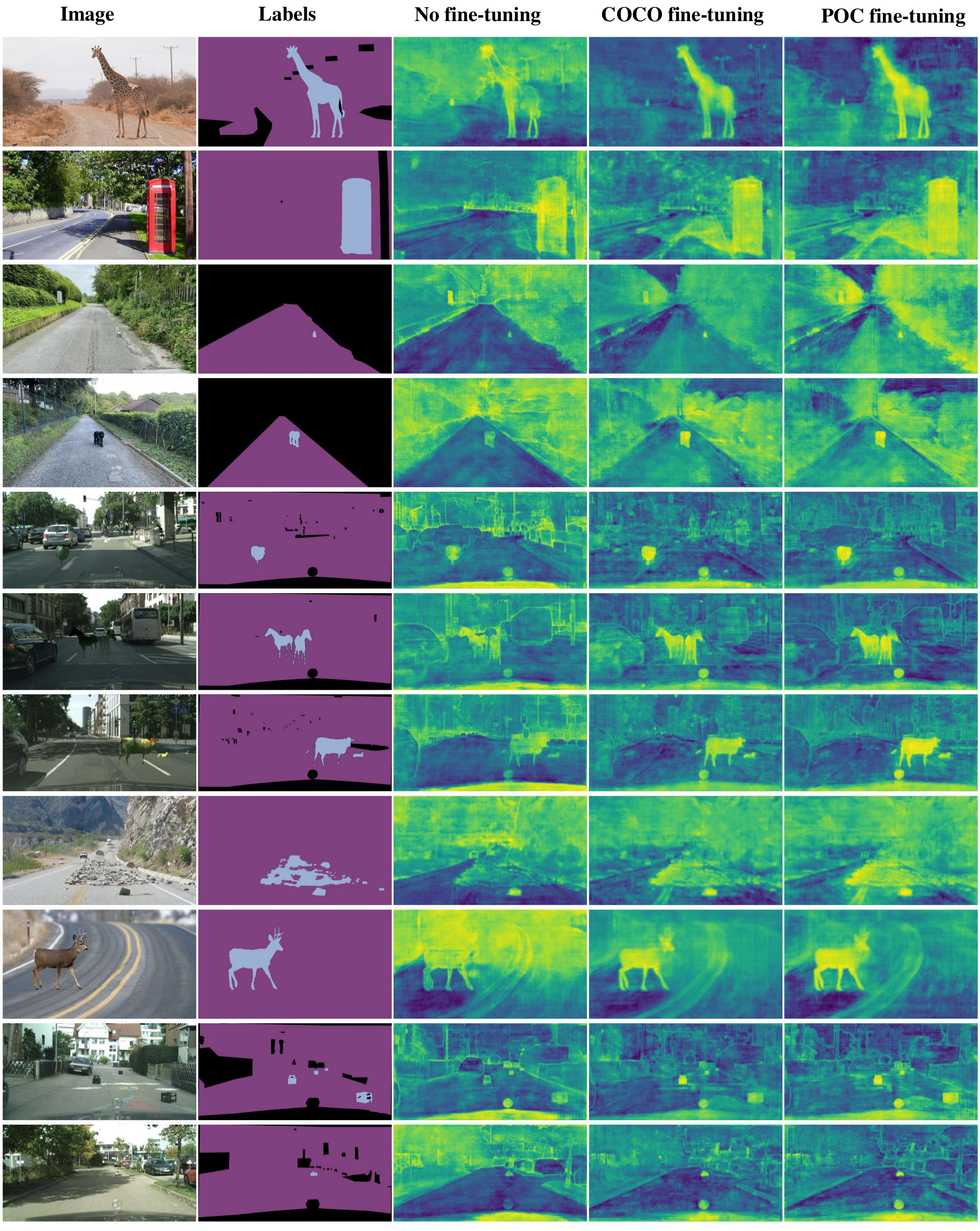}
    \caption{\textbf{RPL anomaly scores on samples from related datasets (see~\cref{fig:dset_comparison})}.}
\end{figure} 

\clearpage

\begin{figure}[ht]
    \centering
    \includegraphics[width=0.85\linewidth]{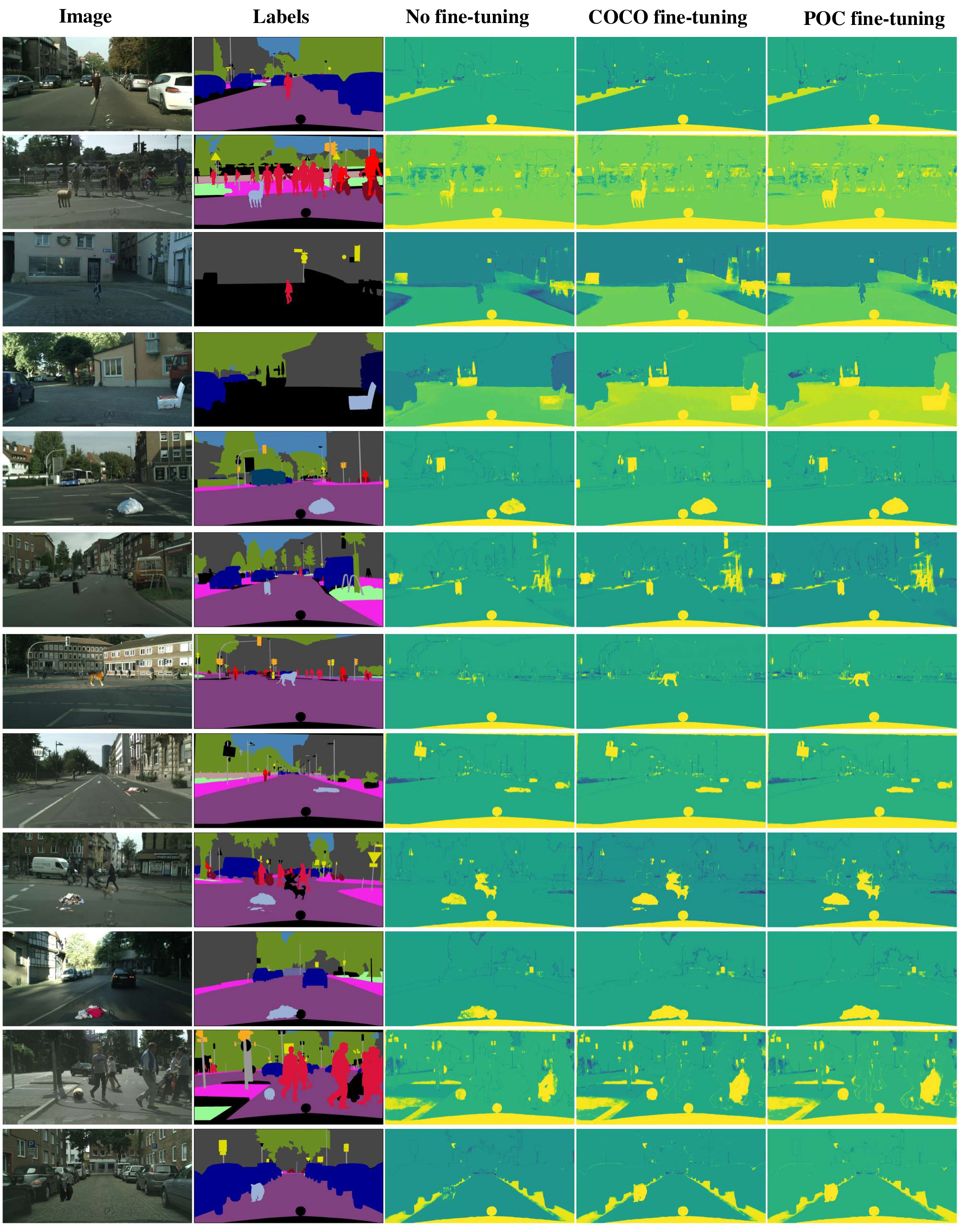}
    \caption{\textbf{RbA anomaly scores on 
    CS-POC 
    samples.}}
\end{figure} 

\clearpage

\begin{figure}[ht]
    \centering
    \includegraphics[width=0.85\linewidth]{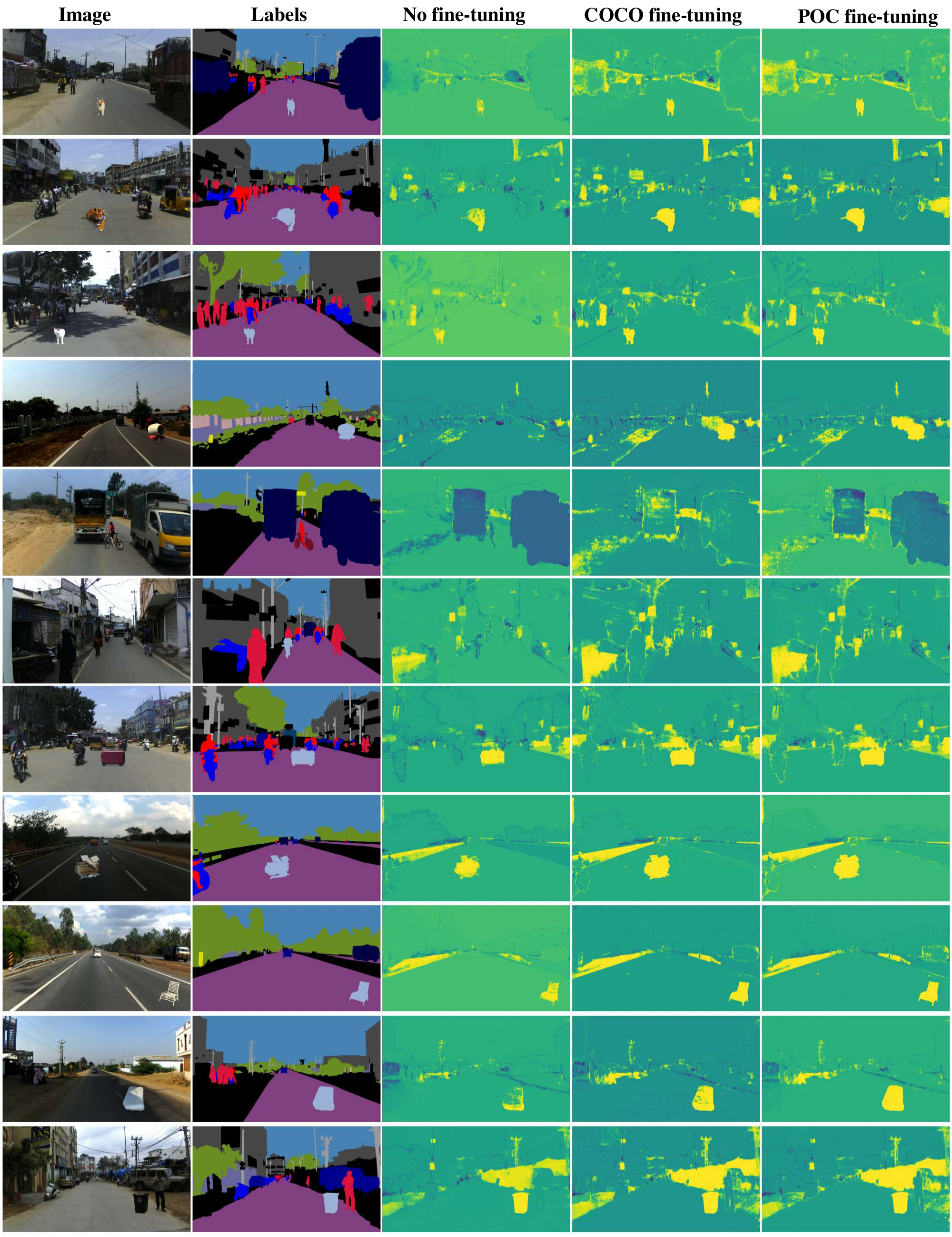}
    \caption{\textbf{RbA anomaly scores on IDD-POC 
    samples.}}
\end{figure} 

\clearpage

\begin{figure}[ht]
    \centering
    \includegraphics[width=0.85\linewidth]{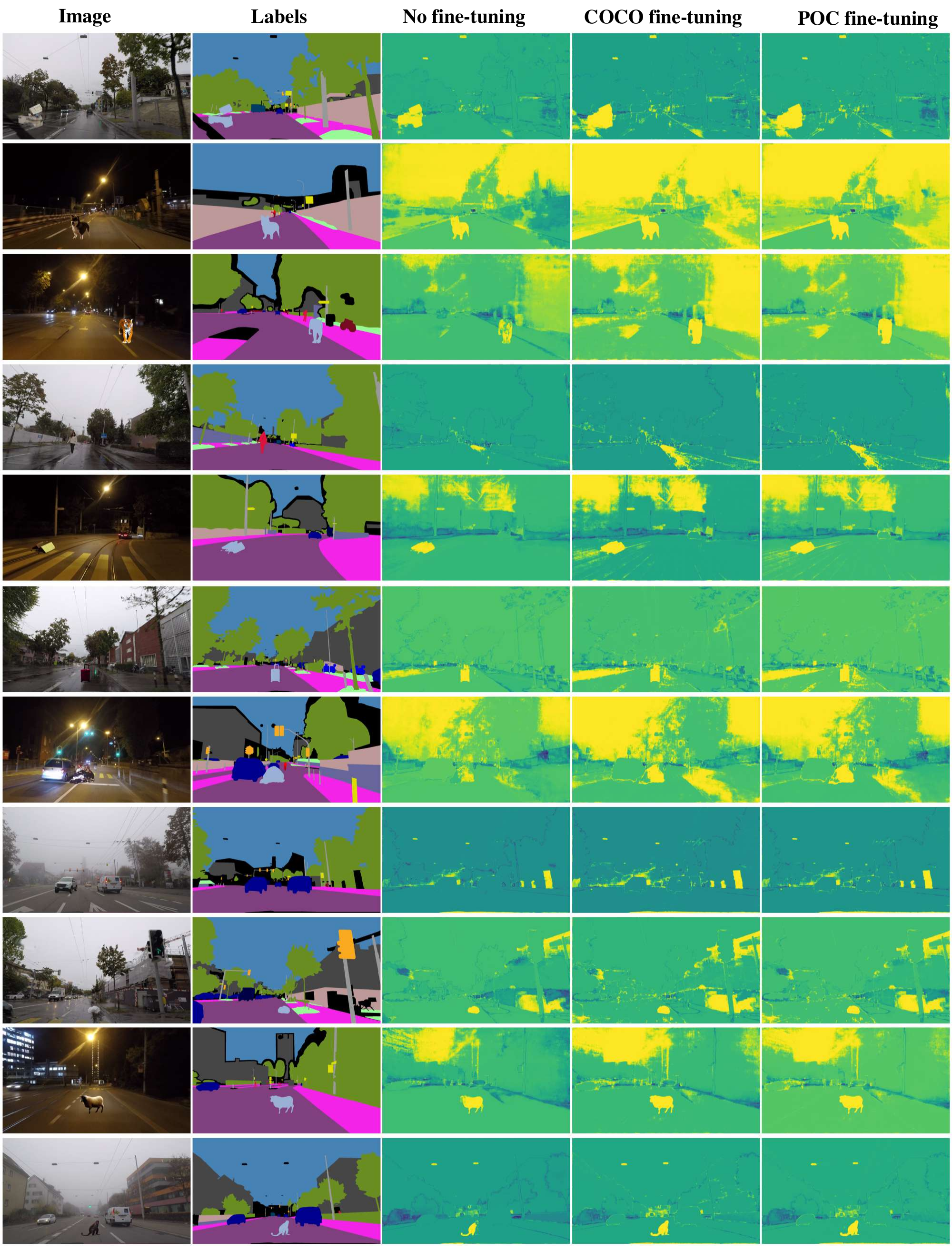}
    \caption{\textbf{RbA anomaly scores on ACDC-POC samples.}}
\end{figure} 

\clearpage

\begin{figure}[ht]
    \centering
    \includegraphics[width=0.85\linewidth]{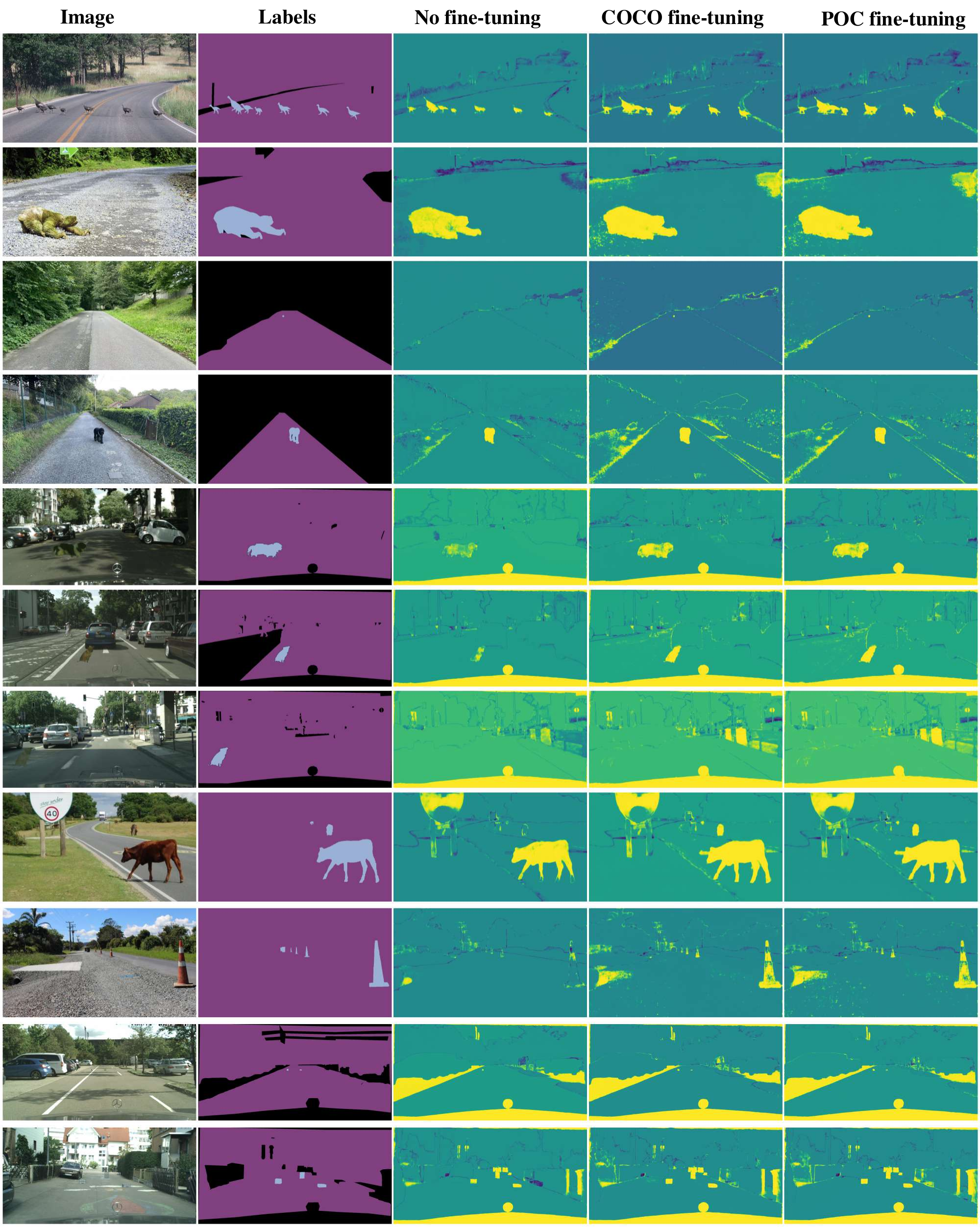}
    \caption{\textbf{RbA anomaly scores on samples from related datasets (see~\cref{fig:dset_comparison})}.}
\end{figure} 

\clearpage

\section{Additional qualitative results}
\label{sec:web_images_preds}
In this section we show additional qualitative results for the dataset extension experiments (\textit{c.f.} \cref{sec:dataset_extension}). We show predictions on all evaluated datasets for DLV3+, ConvNeXt and Segmenter models.

\begin{figure}[ht]
    \centering
    \includegraphics[width=0.55\linewidth]{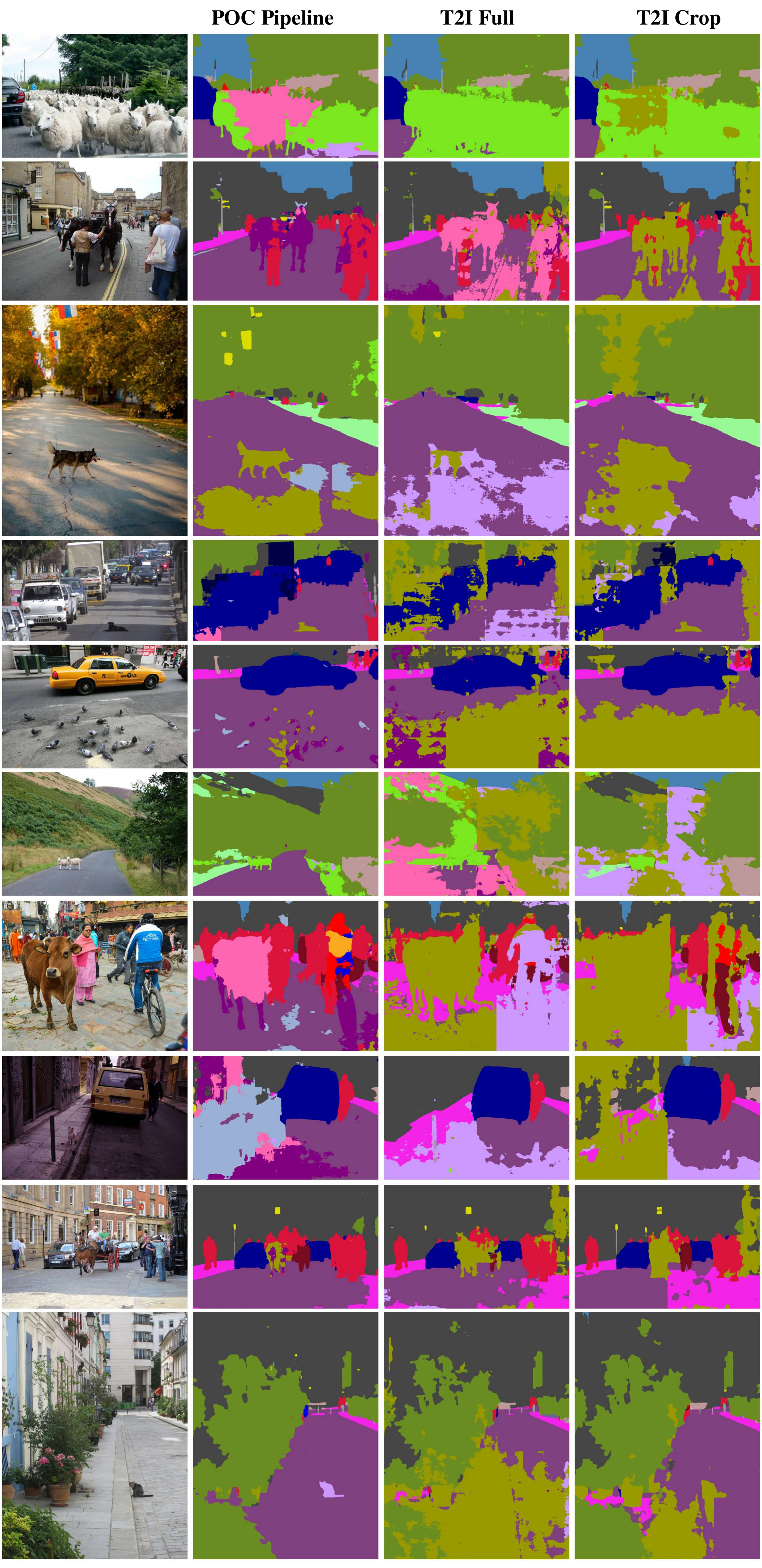}
    \caption{\textbf{DLV3+ predictions on additional web images.}}
\end{figure} 

\clearpage

\begin{figure}[ht]
    \centering
    \includegraphics[width=0.55\linewidth]{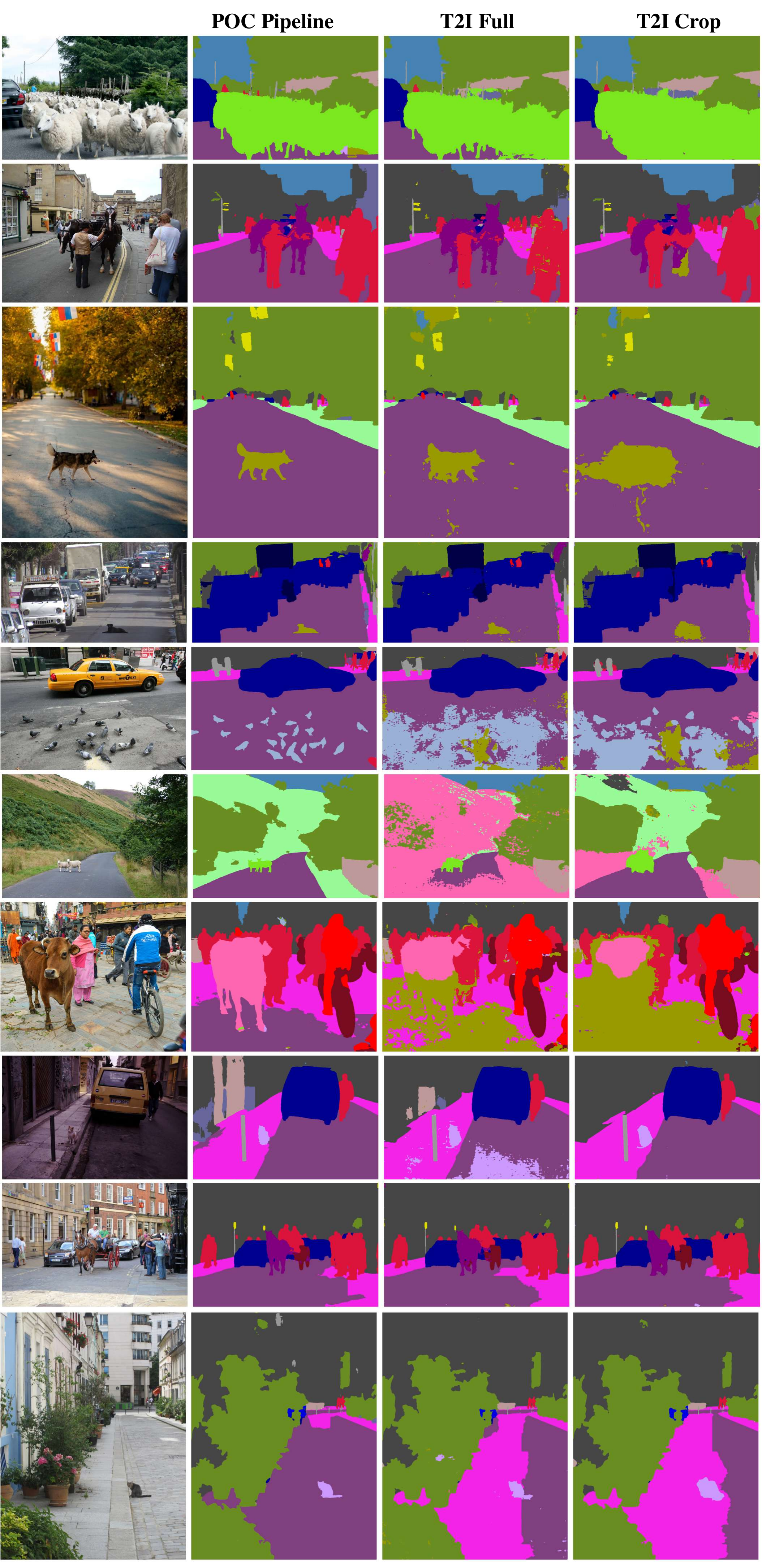}
    \caption{\textbf{ConvNeXt predictions on additional web images.}}
\end{figure} 

\clearpage

\begin{figure}[ht]
    \centering
    \includegraphics[width=0.55\linewidth]{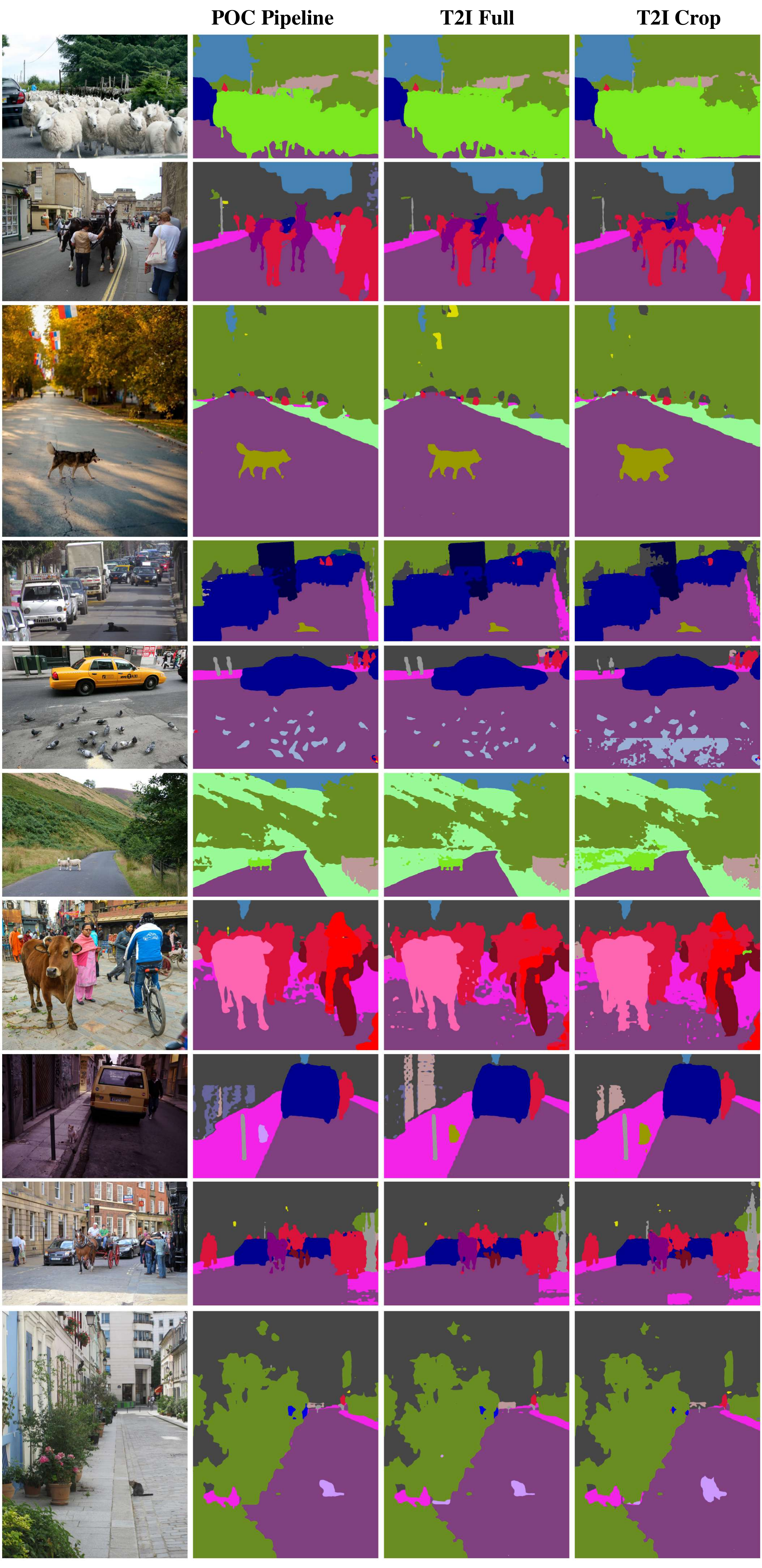}
    \caption{\textbf{Segmenter predictions on additional web images.}}
\end{figure} 

\clearpage

\begin{figure}[ht]
    \centering
    \includegraphics[width=0.85\linewidth]{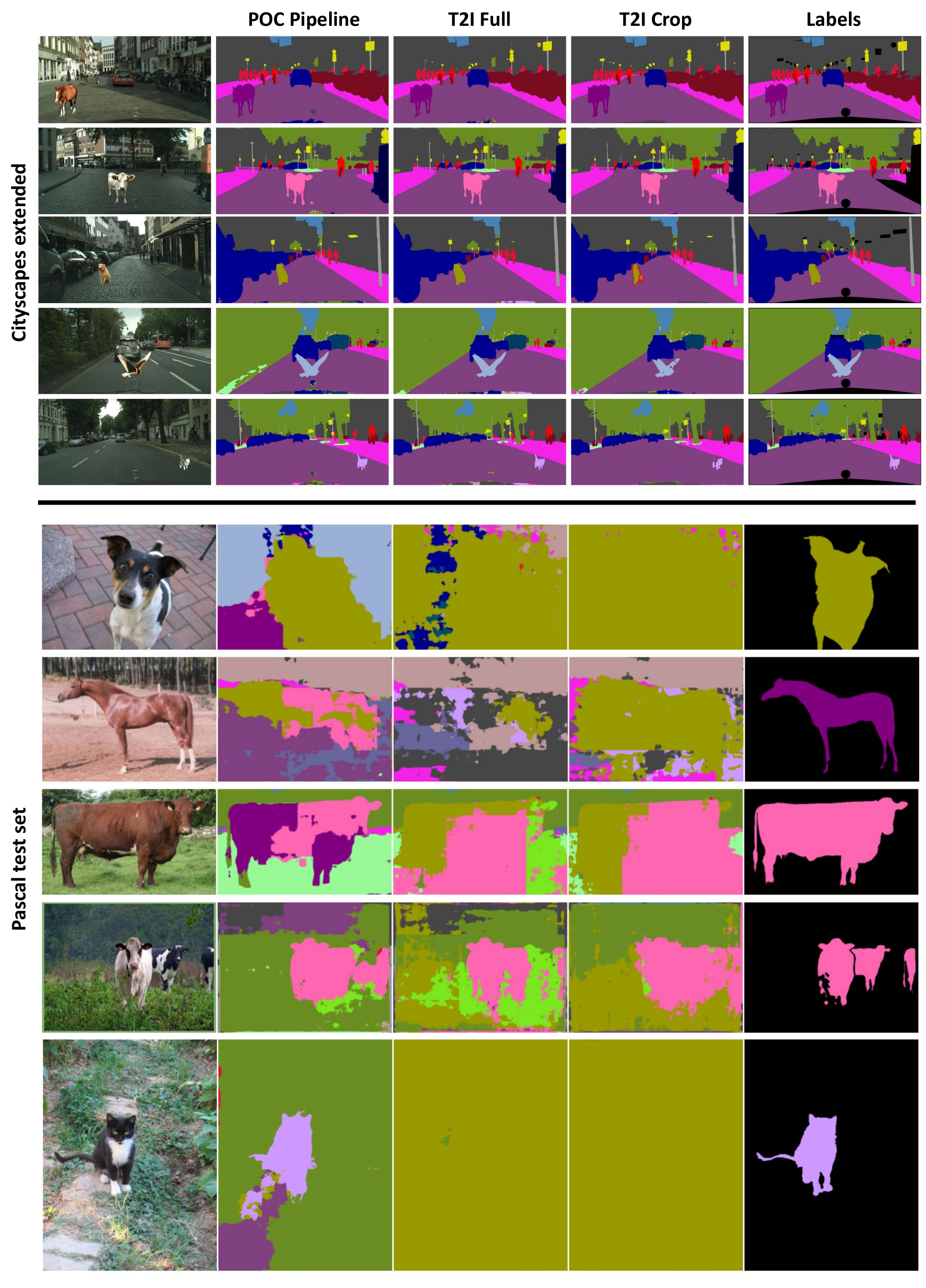}
    \caption{\textbf{DLV3+ predictions on extended Cityscapes (\textit{POC A}) and Pascal validation sets.}}
\end{figure} 

\clearpage

\begin{figure}[ht]
    \centering
    \includegraphics[width=0.85\linewidth]{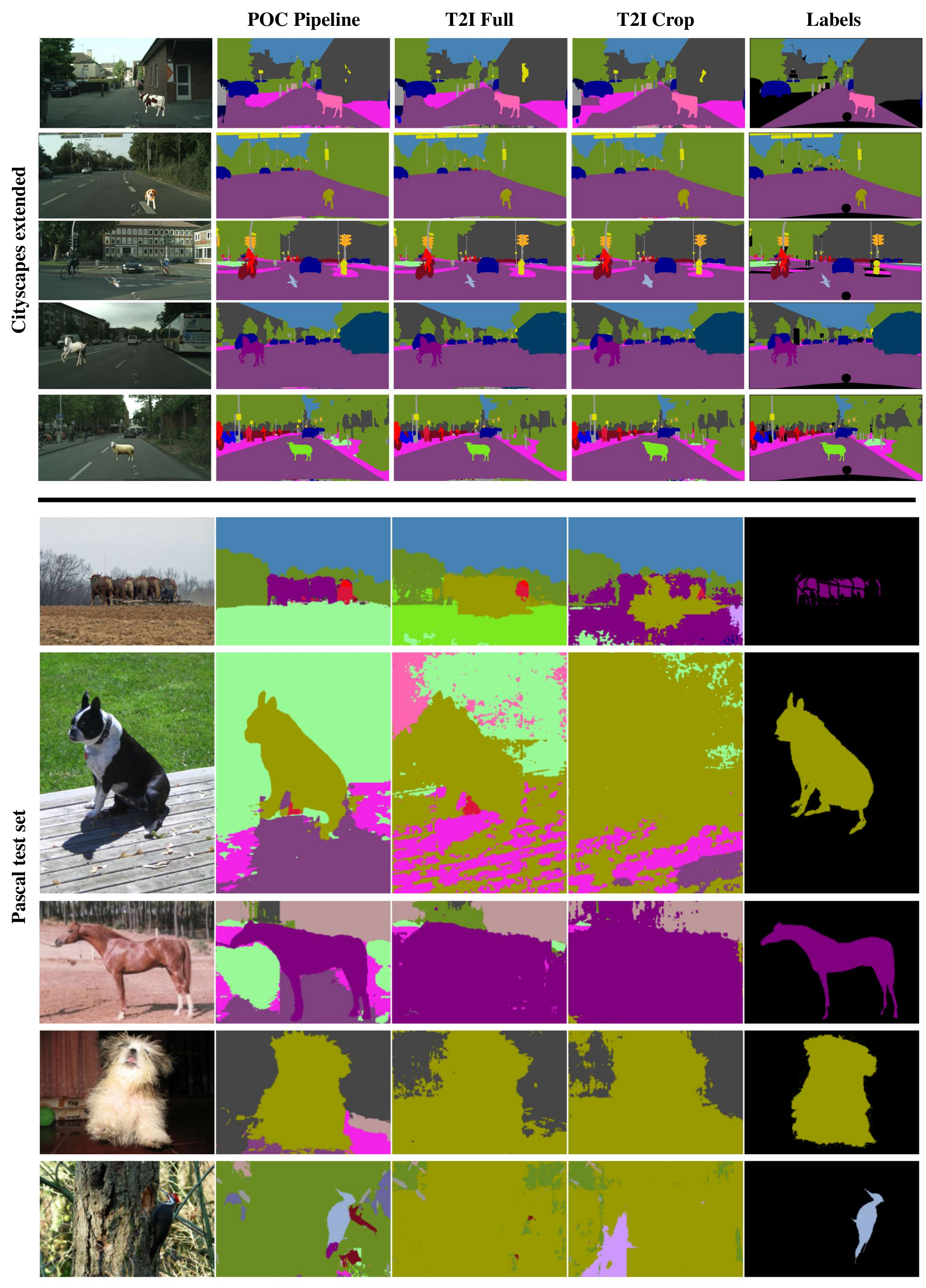}
    \caption{\textbf{ConvNeXt predictions on extended Cityscapes (\textit{POC A}) and Pascal validation sets.}}
\end{figure} 

\clearpage

\begin{figure}[ht]
    \centering
    \includegraphics[width=0.85\linewidth]{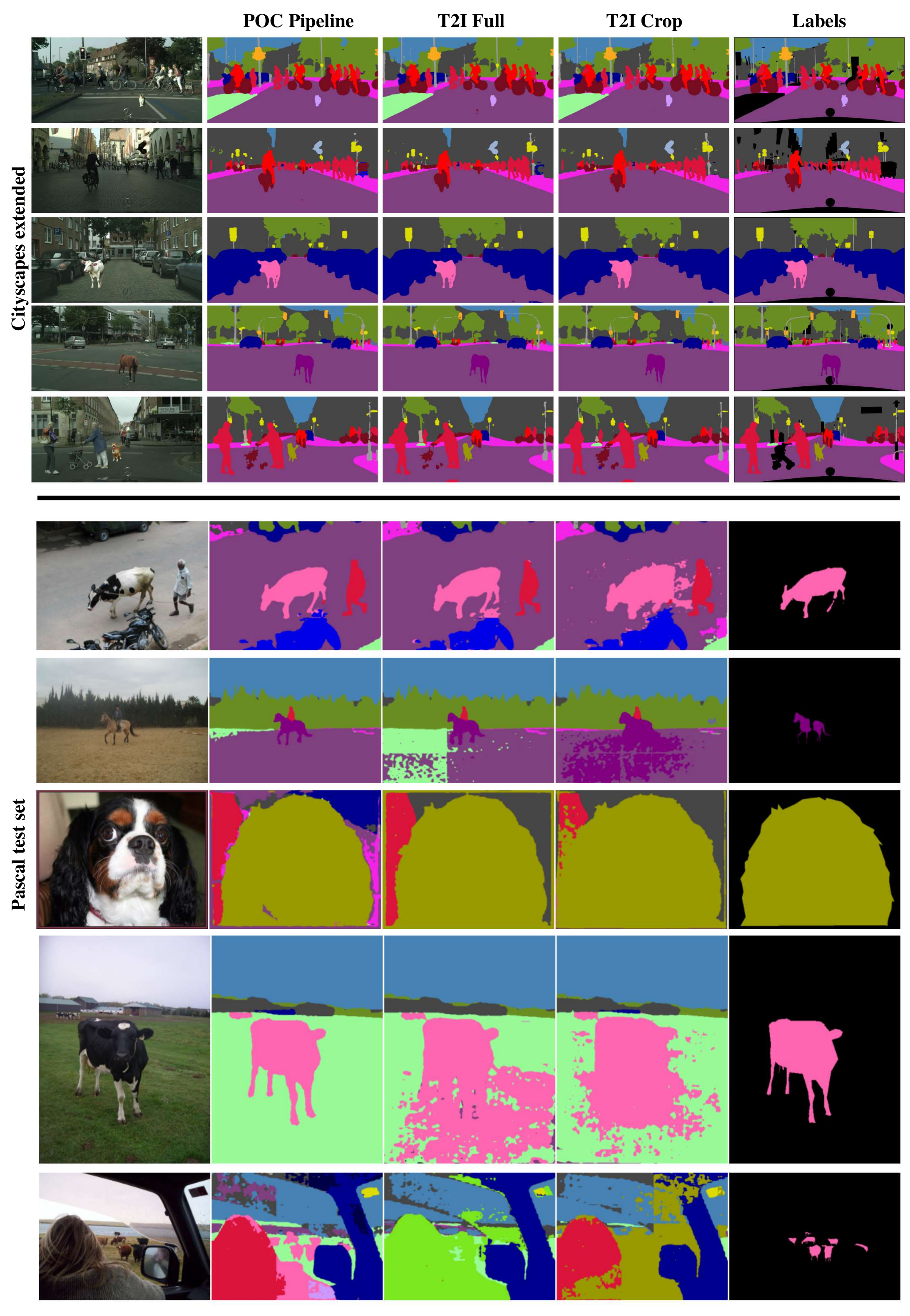}
    \caption{\textbf{Segmenter predictions on extended Cityscapes (\textit{POC A}) and Pascal validation sets.}}
\end{figure} 

\clearpage

\end{document}